\documentclass{article}

\usepackage[noruler,preprint]{iclr2027_conference}
\usepackage{times}
\usepackage{amsmath,amssymb,booktabs,graphicx,microtype,float,wrapfig,flafter,placeins}
\usepackage{xcolor,array,multirow,enumitem}
\usepackage{colortbl}  % Tianci: cell tints in the main results table
\usepackage{pifont}  % Tianci: check/cross marks for the benchmark comparison table
\newcommand{\benchname}{\textsc{NetBreaker}}
\newcommand{\cmark}{\ding{51}}
\newcommand{\xmark}{\textcolor{gray!70}{\ding{55}}}

\usepackage{hyperref}
\hypersetup{hidelinks}

\newcommand{\Tianci}[1]{#1}
\newcommand{\Wenchuan}[1]{#1}

\title{\Tianci{NetBreaker: Benchmarking Network Dismantling across Objectives, Scales, and Topologies}}

\author{%
  Wenchuan Yang\textsuperscript{1}, Tianci Bu\textsuperscript{2},
  Suoyi Tan\textsuperscript{1}, Changjun Fan\textsuperscript{1},
  Zijie Zhou\textsuperscript{2}, Xin Lu\textsuperscript{1,*}\\[3pt]
  \normalfont\textsuperscript{1} National University of Defense Technology\\
  \normalfont\textsuperscript{2} The Hong Kong University of Science and Technology%
}
\hypersetup{%
  pdftitle={NetBreaker: Benchmarking Network Dismantling across Objectives, Scales, and Topologies},
  pdfauthor={Wenchuan Yang, Tianci Bu, Suoyi Tan, Changjun Fan, Zijie Zhou, Xin Lu}%
}

\newcommand{\rhoT}{\rho_{0.10}}
\newcommand{\aucT}{\mathrm{AUC}_{0.10}}
\newcommand{\keyinsights}[2]{%
  \par\medskip
  \noindent\begingroup
  \setlength{\fboxrule}{0.6pt}%
  \setlength{\fboxsep}{6pt}%
  \fbox{\begin{minipage}{\dimexpr\linewidth-2\fboxsep-2\fboxrule\relax}
    \textcolor{blue!55!black}{\textbf{Key Insights for #1:}}\if\relax\detokenize{#2}\relax\else\ #2\fi\strut
  \end{minipage}}%
  \endgroup
  \par\medskip
}

\begin{document}
\maketitle
\begingroup
\renewcommand{\thefootnote}{*}
\footnotetext{Corresponding author: Xin Lu
  (\href{mailto:xin.lu.lab@outlook.com}{xin.lu.lab@outlook.com}).
  Wenchuan Yang: \href{mailto:wenchuanyang97@163.com}{wenchuanyang97@163.com}.}
\endgroup

\begin{abstract}
% Tianci: abstract shortened to ~190 words for the page budget.
\Wenchuan{Network dismantling seeks to fragment a network with as few node removals as possible, which is widely applied in vulnerability analysis, epidemic containment, and infrastructure protection. Although learning-based methods have become increasingly prominent, their progress remains difficult to assess because existing studies differ in method coverage, datasets, evaluation criteria, and experimental protocols. Thus, we introduce \benchname{}, a comprehensive benchmark designed for the learning era of network dismantling. \benchname{} organizes 30 methods (18 non-learning and 12 learning-based) under a common taxonomy and evaluates them on 185 networks from eight unified domains, ranging from 14 to 1.69 million nodes. To enable fair comparison, heterogeneous method outputs are converted into node removal sequences and assessed under a unified execution protocol. The evaluation covers four complementary dimensions: target-based effectiveness, trajectory-based effectiveness, structural diagnostics, and applicability \& computational efficiency. Extensive experiments show that current learning-based methods do not consistently outperform strong non-learning approaches, while exposing substantial differences in effectiveness, scalability, and efficiency across network conditions and evaluation perspectives. \benchname{} provides a reproducible basis for measuring progress and developing reliable, scalable, and computationally efficient network dismantling methods.}
\end{abstract}

\section{Introduction}
Complex networks underpin many natural and engineered systems, where interconnectivity supports collective function but can also amplify the impact of disruptions. A network may remain largely intact under random failures yet become highly vulnerable when perturbations are strategically targeted, making such structural fragility a central concern in network robustness \citep{artime2024robustness}. \textbf{Network dismantling (ND)} studies this vulnerability by examining how a small targeted set of nodes can be removed to progressively break network connectivity \citep{braunstein2016network}. Beyond characterizing how susceptible a network is to deliberate disruption, dismantling also provides a basis for reasoning about where connectivity should be protected or intentionally interrupted \citep{fan2020finder,artime2024robustness}. Efficient network dismantling is intrinsically difficult because identifying the smallest set of removals that achieves a prescribed level of fragmentation is an NP-hard combinatorial optimization problem \citep{braunstein2016network}. The challenge is further compounded in real-world networks, whose heterogeneous and multiscale structures make the effectiveness of a removal strategy strongly dependent on topology \citep{wandelt2018comparative,artime2024robustness,adler2026lgdna}.

To tackle these challenges, a wide range of methods have been proposed. Early studies relied mainly on structural centrality measures, ranking nodes according to their individual topological importance \citep{wandelt2018comparative}. As it became clear that effective dismantling depends on the joint effect of multiple removals rather than on isolated node scores, the focus shifted toward optimal-percolation, decycling, message-passing, and graph-partitioning formulations that explicitly account for collective structural effects \citep{morone2015collective,braunstein2016network,mugisha2016bpd,zdeborova2016fast,ren2019gnd}. More recently, machine learning has provided a different way to capture these dependencies. GDM learns graph representations from optimally solved small instances, while FINDER uses deep reinforcement learning to acquire sequential removal policies that can be transferred to unseen networks \citep{grassia2021gdm,fan2020finder}. Subsequent work has further developed supervised, unsupervised, and reinforcement-learning approaches for transferable dismantling \citep{zhang2022nirm,zhang2023dcrs,tian2026mind,zhang2026stable}.

% \begin{wrapfigure}{L}{0.49\linewidth}
%   \vspace{-0.8\baselineskip}
%   \centering
%   \includegraphics[width=\linewidth]{Fig_Intro.png}
%   \caption{Test frequency of 185 datasets.}
%   \label{fig:intro-frequency}
%   \vspace{-0.5\baselineskip}
% \end{wrapfigure}

\begin{figure}[!t]
\centering
\includegraphics[width=0.95\linewidth]{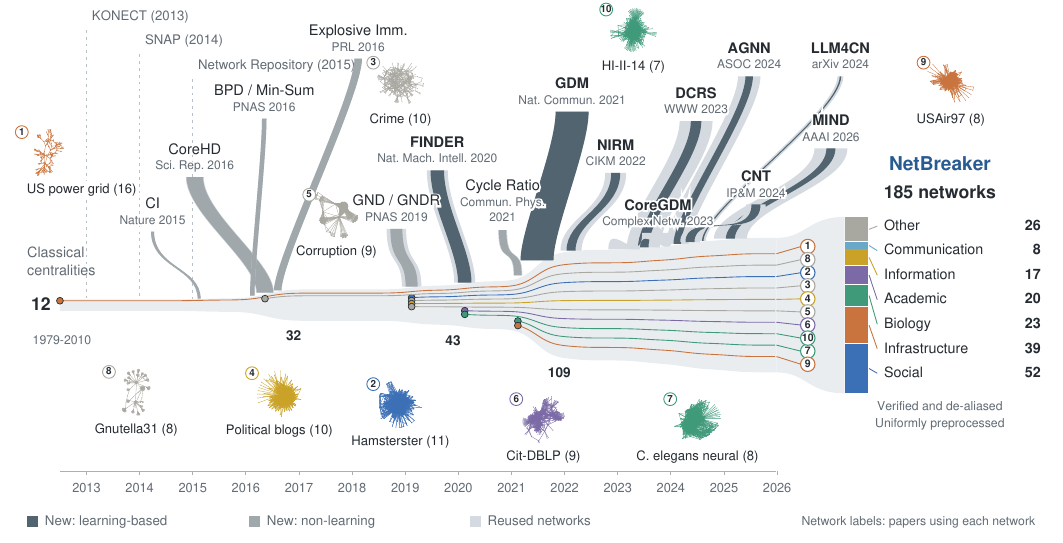}
\caption{Evolution and important network datasets in network dismantling.}
\label{fig:river}
\end{figure}

Despite the rapid development of network dismantling methods, the field still lacks a sufficiently unified basis for assessing progress across studies. This inconsistency arises mainly from three aspects. First, network preparation differs substantially across studies. Methods are evaluated on different collections of synthetic and real-world networks, often with different graph construction and preprocessing procedures, making results difficult to compare directly. Second, the dismantling problem itself is not formulated uniformly. Studies may adopt different dismantling targets, stopping conditions, and assumptions about how removals are performed, so nominally similar experiments can correspond to different optimization problems. Third, evaluation protocols remain inconsistent. Different metrics are used to summarize dismantling performance, while computational cost and applicability are not always reported under comparable conditions. Although previous comparative studies have provided valuable assessments of representative methods \citep{wandelt2018comparative,artime2024robustness} and recent work has expanded evaluation to large collections of real-world networks \citep{adler2026lgdna}, these differences continue to hinder a consistent assessment of methodological progress.

To provide a common basis for comparison, we introduce \textsc{NetBreaker}, a systematic benchmark for network dismantling. \textsc{NetBreaker} standardizes network preparation and problem instantiation across methods, while a method-independent evaluation procedure assesses the resulting removal sequences under consistent criteria. The benchmark covers 30 dismantling methods and variants across 185 real-world networks. By separating method execution from evaluation, \textsc{NetBreaker} preserves the original behavior of heterogeneous algorithms while enabling their performance, computational cost, and execution coverage to be examined under the same experimental setting.

Our contributions can be summarized as follows:
\begin{itemize}[leftmargin=1.35em,labelsep=0.45em]
\item \textbf{Comprehensive Benchmark for the Learning Era of Network Dismantling.} NetBreaker enables a fair and direct comparison among 18 non-learning and 12 learning-based network dismantling methods and variants by unifying their experimental settings across 185 networks with diverse domains, scales, and topological characteristics. By bringing traditional algorithms and recently developed learning-based approaches into the same benchmark, NetBreaker provides a comprehensive view of the methodological progress in network dismantling.

\item \textbf{Multi-faceted Evaluation and Analysis.} We systematically evaluate existing network dismantling methods from multiple complementary dimensions, including dismantling effectiveness, computational efficiency, completion coverage, scalability, and performance across different network characteristics. Our analysis reveals the respective strengths, limitations, and applicable conditions of different families of methods, providing practical insights to guide future research.

\item \textbf{A Unified and Resource-Aware Benchmarking Codebase.} We develop a unified and extensible codebase that accommodates the heterogeneous execution procedures of existing network dismantling methods. In addition to solution quality, the codebase is able to record computational costs and method completion under predefined runtime, memory, and scale constraints. It provides a reproducible foundation for evaluating existing methods and integrating future network dismantling approaches.
\end{itemize}

\section{Problem and Positioning}
\label{sec:task}

\Wenchuan{\textbf{Task.} We study node dismantling on a simple, undirected, and unweighted graph
$G=(V,E)$ with $N=|V|$ nodes. Removing a set $S\subseteq V$ yields the
residual graph $G[V\setminus S]$. Given a fragmentation threshold
$\theta\in(0,1)$, the objective is to find the smallest removal set whose
deletion reduces the largest connected component to at most a
$\theta$ fraction of the original network:
\[
S^{*}
=
\arg\min_{S\subseteq V}|S|
\quad
\mathrm{s.t.}
\quad
\frac{\left|\mathrm{LCC}\!\left(G[V\setminus S]\right)\right|}{N}
\leq \theta.
\]
}
% Tianci: Section 2.2 was two pages of prose that restated the introduction. It is replaced by
% one positioning paragraph and a comparison table, which is what reviewers of benchmark papers
% look for first. Cells for prior work are filled from the papers and our registry; cells marked
% [v] must be re-verified against the original papers before submission.
\Wenchuan{\textbf{Why a new benchmark.}
Existing evaluations of network dismantling methods broadly fall into two groups. Comparative studies evaluate multiple methods under shared experimental settings. Wandelt et al.\ \citep{wandelt2018comparative} compare 13 predominantly non-learning methods on 20 networks, assessing dismantling performance through the robustness measure $R$ over the LCC trajectory together with execution time. Artime et al.\ \citep{artime2024robustness} compare 13 methods on 65 networks using a 10\% LCC target and the corresponding trajectory AUC. Although these studies provide valuable common test basis, the performance evaluation is conducted by a limited set of perspectives and coverlittle learning-based dismantling methods.}

\Wenchuan{Method-centric studies constitute the second group. To substantiate the effectiveness of a proposed method, these studies often employ broader experimental designs tailored to that method. GDM \citep{grassia2021gdm}, for example, considers multiple evaluation perspectives to validate its dismantling performance, whereas LGD-NA \citep{adler2026lgdna} strengthens the empirical effectiveness for its proposed method by comparing 17 methods across 1,475 networks. These experiments provide informative evidence for the corresponding methods, but their baselines, network collections, and experimental protocols are selected around different research questions. Their results therefore cannot be compared directly across studies and do not constitute a shared benchmark spanning non-learning and learning-based methods.}

\Wenchuan{Besides, the datasets used in both groups of studies are organized under inconsistent domain classifications, and the complete network collections required to reproduce the reported comparisons are not always available. These inconsistencies further affect the direct comparison across studies.}

\Wenchuan{\textsc{NetBreaker} is designed to provide a unified benchmark for the learning era of network dismantling. It evaluates 18 non-learning methods together with 12 recent learning-based methods under the same experimental protocol. Drawing from the networks used in prior studies, the benchmark selects 185 datasets covering a broad range of network scales and application domains, and reorganizes their inconsistent domain labels into a unified domain taxonomy. It further establishes an explicit taxonomy of dismantling methods, provides the complete benchmark dataset collection, and evaluates methodological progress through four complementary dimensions. By integrating these perspectives within a comprehensive evaluation framework,  \textsc{NetBreaker} enables fair and direct comparison across methodological generations.}

\begin{table}[t]
\caption{\Wenchuan{Comparison with related network dismantling study. Domains report dataset-level coverage under the unified domain taxonomy. Evaluation dimensions count target-based evaluation, trajectory-based evaluation, structural diagnostics, and applicability and computational efficiency. For dataset availability, \cmark{} denotes that the exact experimental graphs are provided, Partial that only a subset or acquisition instructions are provided, and \xmark{} that neither is provided.}}
\label{tab:benchmark-comparison}
\centering
\scriptsize
\setlength{\tabcolsep}{3.0pt}
\renewcommand{\arraystretch}{1.08}
\begin{tabular}{@{}lcccccccc@{}}
\toprule
\Wenchuan{Study} & \Wenchuan{Benchmark} & \Wenchuan{Methods} & \Wenchuan{Datasets} & \Wenchuan{Domains} & \Wenchuan{Max. $N$} & \Wenchuan{\shortstack{Eval.\\dims.}} & \Wenchuan{\shortstack{Dataset\\availability}} & \Wenchuan{\shortstack{Method\\taxonomy}} \\
\midrule
\Wenchuan{Wandelt et al.\ \citep{wandelt2018comparative}} & \Wenchuan{\cmark} & \Wenchuan{13} & \Wenchuan{20} & \Wenchuan{6/8} & \Wenchuan{$8.1\times10^{4}$} & \Wenchuan{2} & \Wenchuan{Partial} & \Wenchuan{\cmark} \\
\Wenchuan{Artime et al.\ \citep{artime2024robustness}} & \Wenchuan{\cmark} & \Wenchuan{13} & \Wenchuan{65} & \Wenchuan{8/8} & \Wenchuan{$1.4\times10^{6}$} & \Wenchuan{2} & \Wenchuan{\cmark} & \Wenchuan{\cmark} \\
\Wenchuan{FINDER \citep{fan2020finder}} & \Wenchuan{\xmark} & \Wenchuan{11} & \Wenchuan{9} & \Wenchuan{4/8} & \Wenchuan{$1.6\times10^{6}$} & \Wenchuan{3} & \Wenchuan{\cmark} & \Wenchuan{\cmark} \\
\Wenchuan{GDM \citep{grassia2021gdm}} & \Wenchuan{\xmark} & \Wenchuan{15} & \Wenchuan{57} & \Wenchuan{8/8} & \Wenchuan{$1.8\times10^{6}$} & \Wenchuan{4} & \Wenchuan{Partial} & \Wenchuan{\xmark} \\
\Wenchuan{MIND \citep{tian2026mind}} & \Wenchuan{\xmark} & \Wenchuan{9} & \Wenchuan{48} & \Wenchuan{7/8} & \Wenchuan{$1.5\times10^{6}$} & \Wenchuan{2} & \Wenchuan{\cmark} & \Wenchuan{\cmark} \\
\Wenchuan{LGD-NA \citep{adler2026lgdna}} & \Wenchuan{\xmark} & \Wenchuan{17} & \Wenchuan{1,475} & \Wenchuan{8/8} & \Wenchuan{$2.3\times10^{4}$} & \Wenchuan{3} & \Wenchuan{Partial} & \Wenchuan{\cmark} \\
\midrule
\Wenchuan{\benchname{} (this work)} & \Wenchuan{\cmark} & \Wenchuan{30} & \Wenchuan{185} & \Wenchuan{8/8} & \Wenchuan{$1.7\times10^{6}$} & \Wenchuan{4} & \Wenchuan{\cmark} & \Wenchuan{\cmark} \\
\bottomrule
\end{tabular}
\end{table}

\section{The \benchname{} Benchmark}
\label{sec:benchmark}

\begin{figure}[t]
\centering
\includegraphics[width=\linewidth]{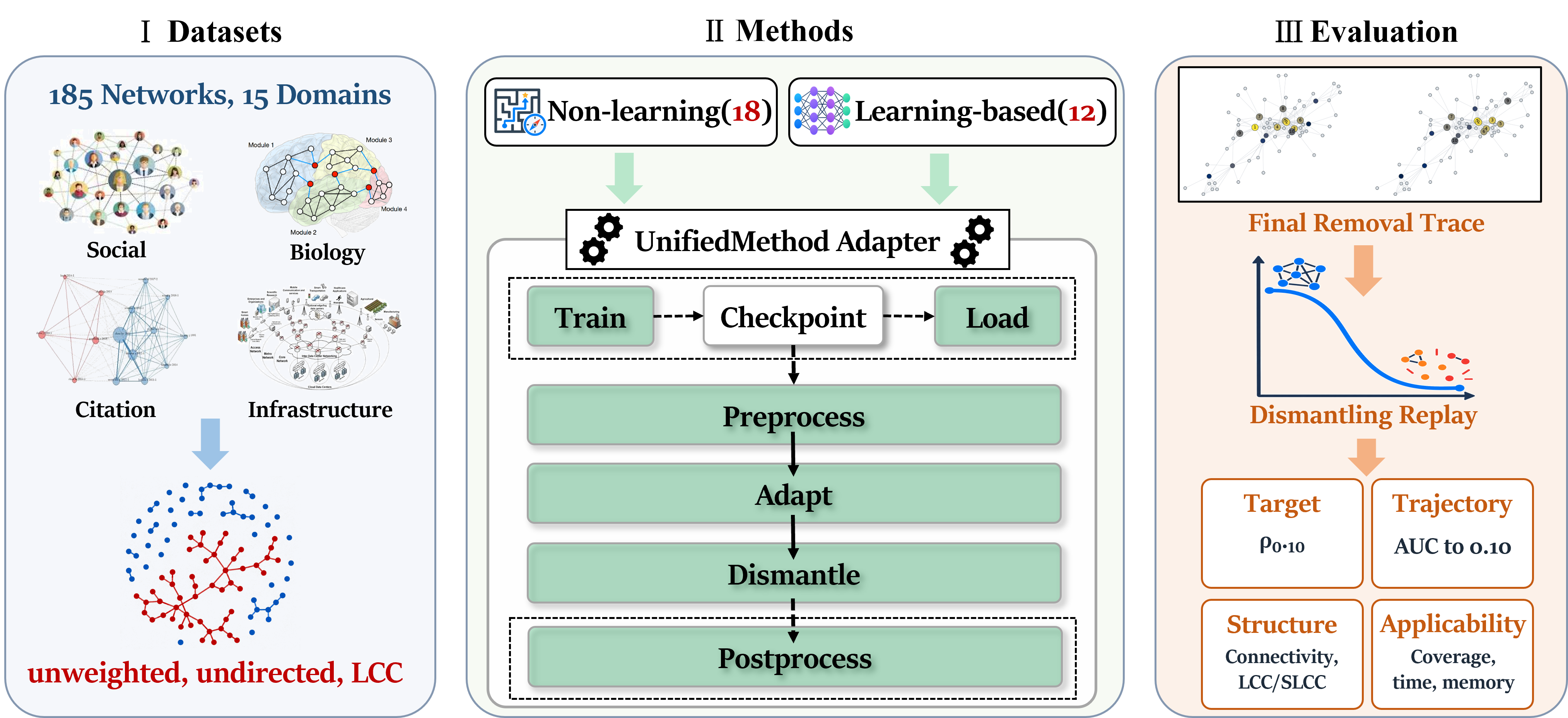}
\caption{\Wenchuan{\benchname{} at a glance. Our benchmark mainly consists of Datasets, Methods, and Evaluation modules. Within the Datasets module, we collect 185 network datasets and convert each network into a unified verison. As for Method module, we design a standard procedure to execute different methods. The output of all the methods are then set to be node removal sequence, which is served as the basis for the comprehensive evaluation.}}
\label{fig:landscape}
\end{figure}

% Tianci (round 3): shortened. The method-family table and the protocol table moved to
% Appendices B and C; Figure 1 now carries the suite and the protocol. Research questions expanded.
\Wenchuan{\benchname{} follows a three-stage procedure (Figure~\ref{fig:landscape}). The first stage constructs a common network corpus that includes networks of different scales and application domains, and all methods operate on the same
canonically prepared network instances. Second, each method retains its method-specific training, optimization, inference, and post-processing procedure, while its output is converted into a common node removal sequence. The third stage evaluates their outputs through a common representation and four complementary evaluation dimensions under a unified execution protocol.}

\subsection{Networks}
\label{sec:networks}
\textbf{Collection and preparation.}
We construct the corpus through a literature-guided collection process, tracing networks evaluated in network dismantling and related critical-node identification studies. The registry covers classical centrality baselines and subsequent non-learning and learning-based methods, including studies published in venues such as Nature, PNAS, Nature Machine Intelligence, Nature Communications, CIKM, and The Web Conference, alongside relevant preprints. We link reported networks to their source studies and retrieve the data from KONECT \citep{kunegis2013konect}, SNAP \citep{leskovec2014snap}, Network Repository \citep{rossi2015network}, LINQS \citep{sen2008collective}, and author-released resources. Figure~\ref{fig:river} illustrates the introduction and reuse of networks across representative studies.
After reconciling dataset identities and provenance across different sources, we retain 185 networks. Each network is represented in the simple, undirected, and unweighted setting defined in Section~\ref{sec:task}, providing identical structural inputs to all evaluated methods. Detailed identity checks and conversion rules are reported in Appendix~\ref{app:datasets}.

\textbf{Coverage and characteristics.}
The processed networks range from 14 to approximately 1.69 million nodes, comprising 52 small ($N<10^3$), 99 medium ($10^3\leq N<5\times10^4$), and 34 large ($N\geq5\times10^4$) networks. They span social, infrastructure, biological, academic, information, communication, and other application settings, organized through a common domain taxonomy documented in Appendix~\ref{app:datasets}. Average degree ranges from approximately 2.1 to 685, average clustering from nearly zero to 0.94, and degree assortativity from $-0.88$ to $0.89$. This coverage supports comparisons across network scales, domains, and structural characteristics; all statistics refer to the processed graphs.

\subsection{Methods}
\label{sec:methods}
\Wenchuan{\textsc{NetBreaker} evaluates 30 methods, comprising 18 non-learning and 12 learning-based methods. Static, adaptive, and reinsertion variants are retained when they correspond to distinct dismantling procedures and produce different removal sequences. The complete method list, taxonomy, implementation sources, and configurations are reported in Appendix~\ref{app:algorithms}.}

\paragraph{\Wenchuan{Non-learning methods.}}
\Wenchuan{The non-learning methods include a random-based baseline, classical centrality heuristics, approximate-theory methods, decycling algorithms, and graph-partitioning or optimization-based approaches. They range from widely used strategies such as Degree, Betweenness, PageRank, and k-core to methods developed specifically for optimal percolation and network dismantling, including Collective Influence, BPD, Min-Sum, CoreHD, GND, Explosive Immunization, and LGD-NA CND \citep{morone2015collective,mugisha2016bpd,braunstein2016network,zdeborova2016fast,ren2019gnd,clusella2016explosive,adler2026lgdna}.}

\paragraph{\Wenchuan{Learning-based methods.}}
\Wenchuan{The learning-based methods cover supervised learning, self- or unsupervised learning, reinforcement learning, and generative program search. Direct dismantling methods include GDM, NIRM, Stable-AGDM, DCRS, FINDER, and MIND, together with their applicable variants \citep{grassia2021gdm,zhang2022nirm,zhang2023dcrs, fan2020finder,tian2026mind,zhang2026stable}. We additionally include learning-based critical-node methods as explicitly identified proxy baselines because they produce node rankings that can be evaluated through the same removal-sequence interface.}

\subsection{\Wenchuan{Evaluation}}
\label{sec:evaluation}

\paragraph{\Wenchuan{Unified Output Representation.}}
\label{sec:procedure}
\Wenchuan{Dismantling methods produce heterogeneous outputs, including node scores, removal sets, and sequential decisions. Comparing them requires a common representation from which network fragmentation can be measured consistently. Since each intervention removes a node, we use the node removal sequence as this interface, preserving each method's solution procedure and post-processing. Scores induce a descending removal order, while set-valued solutions are ordered through the corresponding method's post-processing.}

\Wenchuan{Given a removal sequence $\pi=(v_1,v_2,\ldots)$, the residual network and its normalized largest connected component after $k$ removals are}
\begin{equation}
\Wenchuan{
G_k=G\!\left[V\setminus\{v_1,\ldots,v_k\}\right],
\qquad
L(k)=\frac{|\mathrm{LCC}(G_k)|}{N}.
}
\end{equation}
\Wenchuan{A shared evaluator reconstructs these states on the same processed network for every method.}

\paragraph{\Wenchuan{Multi-faceted Evaluation.}}
\Wenchuan{Our evaluation is designed to assess both the quality of a dismantling solution and the resources required to obtain it. Quality concerns not only how many nodes must be removed to reach the fragmentation target, but also how much connectivity remains during sequential removal. These two summaries can still obscure how connected components split, motivating additional structural diagnostics. Practical use further requires knowing whether a method completes within the execution constraints and at what computational cost. These considerations lead to the following four evaluation dimensions.}

\noindent\Wenchuan{\textbf{Target-based evaluation.}
To measure the intervention required to reach the prescribed fragmentation level, we use the removal fraction at the first step satisfying the target. With $s=\lceil0.10N\rceil$,}
\begin{equation}
\Wenchuan{
K=\min\{k:|\mathrm{LCC}(G_k)|\leq s\},
\qquad
\rhoT=\frac{K}{N}.
}
\end{equation}
\Wenchuan{A smaller $\rhoT$ indicates that fewer removals are required.}

\noindent\Wenchuan{\textbf{Trajectory-based evaluation.}
To account for connectivity retained during dismantling, we measure the area under the normalized LCC trajectory up to the same target:}
\begin{equation}
\Wenchuan{
\aucT=\frac{1}{N}\operatorname{Simpson}
\bigl(L(1),\ldots,L(K);\,\Delta k=1\bigr).
}
\end{equation}
\Wenchuan{A smaller value indicates lower accumulated connectivity before the target is reached.}

\noindent\Wenchuan{\textbf{Structural diagnostics.}
We examine pairwise connectivity and the evolution of the largest and second-largest connected components to characterize how fragmentation occurs. Diagnostics include mean pairwise connectivity up to the target, the largest single-step LCC drop, and the peak size of the second-largest component, together with the removal fractions at which the latter two occur. They describe structural changes along the trajectory rather than define an additional overall ranking.}

\noindent\Wenchuan{\textbf{Applicability and computational efficiency.}
We report completion coverage, defined as the fraction of networks on which a method produces a valid result under the benchmark constraints, and retain unsuccessful outcomes in cross-method comparisons. Computational cost is measured by observed wall-clock time and peak memory consumption, separating offline training from computation on the target network. The released experiment records preserve execution limits, non-completion categories, aggregation rules, and cost boundaries for every run.}

\subsection{Research Questions}
\label{sec:rqs}

\Wenchuan{We design \textsc{NetBreaker} to systematically evaluate existing network dismantling methods from multiple perspectives. In particular, we aim to investigate the following research questions.}

\paragraph{RQ1: How effective are learning-based methods for network dismantling?}\mbox{}\\
\Wenchuan{\textbf{Motivation and Experiment Design.}
Learning-based methods have increasingly been introduced into network dismantling, yet they are commonly evaluated on different network collections and against different sets of baselines. It therefore remains unclear whether these methods provide consistent improvements over strong non-learning approaches under a unified experimental setting. We compare learning-based and non-learning methods across the full benchmark, examining their dismantling performance and completion coverage under the same preprocessing, execution, and evaluation protocol.}

\paragraph{RQ2: How does the performance of network dismantling methods vary across network scales and domains?}\mbox{}\\
\Wenchuan{\textbf{Motivation and Experiment Design.}
Real-world networks differ substantially in scale and application domain, and a method that performs well on one group of networks may not retain its advantage on another. We therefore evaluate all methods across networks of different scales and across the application domains represented in the benchmark. By comparing their performance rankings and completion coverage within each group, we examine whether the relative strengths of different methods remain stable across heterogeneous network data.}

\begin{table*}[t]
\caption{\Wenchuan{Failure-aware aggregate comparison across 185 benchmark networks. Values are medians over successful dataset-level evaluations. Superscripts \textbf{1--3} mark the three best failure-aware average ranks for each metric. The upward arrows on max $\Delta$LCC and max SLCC indicate larger transition magnitudes, not standalone dismantling superiority.}}
\label{tab:main_results_full}
\centering
\scriptsize
\setlength{\tabcolsep}{3.0pt}
\renewcommand{\arraystretch}{0.96}
\resizebox{\textwidth}{!}{%
\begin{tabular}{r l c ccc cccc c}
\toprule
& & \multicolumn{1}{c}{\Wenchuan{Target-based}} & \multicolumn{2}{c}{\Wenchuan{Trajectory-based}} & \multicolumn{4}{c}{\Wenchuan{Transition diagnostics}} & \\
\cmidrule(lr){3-3}\cmidrule(lr){4-5}\cmidrule(lr){6-9}
\Wenchuan{ID} & \Wenchuan{Method} & \Wenchuan{$\rho_{0.10}\downarrow$} & \Wenchuan{LCC-AUC$\downarrow$} & \Wenchuan{PairConn$\downarrow$} & \Wenchuan{max $\Delta$LCC$\uparrow$} & \Wenchuan{$\rho(\Delta)\downarrow$} & \Wenchuan{max SLCC$\uparrow$} & \Wenchuan{$\rho(\mathrm{SLCC})\downarrow$} & \Wenchuan{Coverage} \\
\midrule
\Wenchuan{1} & \Wenchuan{Random} & \Wenchuan{0.773} & \Wenchuan{0.397} & \Wenchuan{0.349} & \Wenchuan{0.017} & \Wenchuan{0.411} & \Wenchuan{0.004} & \Wenchuan{0.471} & \Wenchuan{100.0\%} \\
\addlinespace[1.2pt]
\Wenchuan{2} & \Wenchuan{Static Degree} & \Wenchuan{0.200} & \Wenchuan{0.108} & \Wenchuan{0.377} & \Wenchuan{0.047} & \Wenchuan{0.124} & \Wenchuan{0.032} & \Wenchuan{0.142} & \Wenchuan{100.0\%} \\
\Wenchuan{3} & \Wenchuan{Adaptive Degree} & \Wenchuan{0.161} & \Wenchuan{0.089} & \Wenchuan{0.417} & \Wenchuan{0.053} & \Wenchuan{0.116} & \Wenchuan{0.042} & \Wenchuan{0.135} & \Wenchuan{100.0\%} \\
\Wenchuan{4} & \Wenchuan{Static Betweenness} & \Wenchuan{0.234} & \Wenchuan{0.121} & \Wenchuan{\textbf{0.316}\textsuperscript{\textbf{1}}} & \Wenchuan{0.053} & \Wenchuan{0.148} & \Wenchuan{0.040} & \Wenchuan{0.169} & \Wenchuan{87.0\%} \\
\Wenchuan{5} & \Wenchuan{Adaptive Betweenness} & \Wenchuan{0.103} & \Wenchuan{0.054} & \Wenchuan{0.387} & \Wenchuan{0.193} & \Wenchuan{0.054} & \Wenchuan{0.178} & \Wenchuan{0.058} & \Wenchuan{67.0\%} \\
\Wenchuan{6} & \Wenchuan{PageRank} & \Wenchuan{0.175} & \Wenchuan{0.091} & \Wenchuan{\textbf{0.361}\textsuperscript{\textbf{2}}} & \Wenchuan{0.047} & \Wenchuan{0.119} & \Wenchuan{0.031} & \Wenchuan{0.133} & \Wenchuan{100.0\%} \\
\Wenchuan{7} & \Wenchuan{k-core / k-shell} & \Wenchuan{0.272} & \Wenchuan{0.148} & \Wenchuan{0.377} & \Wenchuan{0.047} & \Wenchuan{0.189} & \Wenchuan{0.035} & \Wenchuan{0.211} & \Wenchuan{100.0\%} \\
\addlinespace[1.2pt]
\Wenchuan{8} & \Wenchuan{Cycle Ratio} & \Wenchuan{0.241} & \Wenchuan{0.128} & \Wenchuan{0.347} & \Wenchuan{0.052} & \Wenchuan{0.140} & \Wenchuan{0.037} & \Wenchuan{0.186} & \Wenchuan{83.8\%} \\
\Wenchuan{9} & \Wenchuan{Collective Influence} & \Wenchuan{0.137} & \Wenchuan{\textbf{0.079}\textsuperscript{\textbf{2}}} & \Wenchuan{0.360} & \Wenchuan{0.052} & \Wenchuan{\textbf{0.080}\textsuperscript{\textbf{3}}} & \Wenchuan{0.044} & \Wenchuan{0.098} & \Wenchuan{95.1\%} \\
\Wenchuan{10} & \Wenchuan{BPD} & \Wenchuan{\textbf{0.107}\textsuperscript{\textbf{1}}} & \Wenchuan{\textbf{0.077}\textsuperscript{\textbf{3}}} & \Wenchuan{0.559} & \Wenchuan{\textbf{0.173}\textsuperscript{\textbf{3}}} & \Wenchuan{\textbf{0.095}\textsuperscript{\textbf{2}}} & \Wenchuan{\textbf{0.129}\textsuperscript{\textbf{1}}} & \Wenchuan{\textbf{0.092}\textsuperscript{\textbf{2}}} & \Wenchuan{98.4\%} \\
\Wenchuan{11} & \Wenchuan{Min-Sum} & \Wenchuan{\textbf{0.109}\textsuperscript{\textbf{3}}} & \Wenchuan{0.083} & \Wenchuan{0.582} & \Wenchuan{0.183} & \Wenchuan{0.100} & \Wenchuan{\textbf{0.096}\textsuperscript{\textbf{3}}} & \Wenchuan{0.101} & \Wenchuan{96.8\%} \\
\Wenchuan{12} & \Wenchuan{CoreHD} & \Wenchuan{\textbf{0.107}\textsuperscript{\textbf{2}}} & \Wenchuan{0.082} & \Wenchuan{0.583} & \Wenchuan{\textbf{0.179}\textsuperscript{\textbf{2}}} & \Wenchuan{0.096} & \Wenchuan{\textbf{0.101}\textsuperscript{\textbf{2}}} & \Wenchuan{\textbf{0.093}\textsuperscript{\textbf{3}}} & \Wenchuan{100.0\%} \\
\Wenchuan{13} & \Wenchuan{GND} & \Wenchuan{0.147} & \Wenchuan{0.071} & \Wenchuan{0.380} & \Wenchuan{0.134} & \Wenchuan{0.071} & \Wenchuan{0.118} & \Wenchuan{0.087} & \Wenchuan{93.0\%} \\
\Wenchuan{14} & \Wenchuan{GNDR} & \Wenchuan{0.117} & \Wenchuan{0.101} & \Wenchuan{0.736} & \Wenchuan{\textbf{0.330}\textsuperscript{\textbf{1}}} & \Wenchuan{0.112} & \Wenchuan{0.095} & \Wenchuan{0.105} & \Wenchuan{93.0\%} \\
\Wenchuan{15} & \Wenchuan{Explosive Immunization} & \Wenchuan{0.119} & \Wenchuan{\textbf{0.065}\textsuperscript{\textbf{1}}} & \Wenchuan{0.379} & \Wenchuan{0.091} & \Wenchuan{\textbf{0.075}\textsuperscript{\textbf{1}}} & \Wenchuan{0.080} & \Wenchuan{\textbf{0.083}\textsuperscript{\textbf{1}}} & \Wenchuan{99.5\%} \\
\Wenchuan{16} & \Wenchuan{Ensemble-GND} & \Wenchuan{0.109} & \Wenchuan{0.064} & \Wenchuan{0.409} & \Wenchuan{0.218} & \Wenchuan{0.066} & \Wenchuan{0.201} & \Wenchuan{0.069} & \Wenchuan{63.8\%} \\
\Wenchuan{17} & \Wenchuan{Ensemble-GNDR} & \Wenchuan{0.099} & \Wenchuan{0.082} & \Wenchuan{0.732} & \Wenchuan{0.331} & \Wenchuan{0.093} & \Wenchuan{0.096} & \Wenchuan{0.092} & \Wenchuan{63.8\%} \\
\Wenchuan{18} & \Wenchuan{LGD-NA CND} & \Wenchuan{0.168} & \Wenchuan{0.093} & \Wenchuan{0.396} & \Wenchuan{0.080} & \Wenchuan{0.113} & \Wenchuan{0.056} & \Wenchuan{0.118} & \Wenchuan{78.9\%} \\
\addlinespace[1.2pt]
\Wenchuan{19} & \Wenchuan{GDM} & \Wenchuan{0.161} & \Wenchuan{0.095} & \Wenchuan{0.402} & \Wenchuan{0.058} & \Wenchuan{0.121} & \Wenchuan{0.028} & \Wenchuan{0.133} & \Wenchuan{91.9\%} \\
\Wenchuan{20} & \Wenchuan{GDM+R} & \Wenchuan{0.122} & \Wenchuan{0.082} & \Wenchuan{0.485} & \Wenchuan{0.148} & \Wenchuan{0.108} & \Wenchuan{0.095} & \Wenchuan{0.108} & \Wenchuan{90.3\%} \\
\Wenchuan{21} & \Wenchuan{CoreGDM} & \Wenchuan{0.120} & \Wenchuan{0.081} & \Wenchuan{0.508} & \Wenchuan{0.155} & \Wenchuan{0.103} & \Wenchuan{0.098} & \Wenchuan{0.104} & \Wenchuan{91.4\%} \\
\Wenchuan{22} & \Wenchuan{NIRM} & \Wenchuan{0.298} & \Wenchuan{0.134} & \Wenchuan{0.362} & \Wenchuan{0.055} & \Wenchuan{0.140} & \Wenchuan{0.042} & \Wenchuan{0.203} & \Wenchuan{82.2\%} \\
\addlinespace[1.2pt]
\Wenchuan{23} & \Wenchuan{Stable-AGDM} & \Wenchuan{0.195} & \Wenchuan{0.114} & \Wenchuan{0.386} & \Wenchuan{0.056} & \Wenchuan{0.117} & \Wenchuan{0.031} & \Wenchuan{0.142} & \Wenchuan{99.5\%} \\
\Wenchuan{24} & \Wenchuan{Stable-AGDM+R} & \Wenchuan{0.114} & \Wenchuan{0.087} & \Wenchuan{0.556} & \Wenchuan{0.176} & \Wenchuan{0.110} & \Wenchuan{0.096} & \Wenchuan{0.105} & \Wenchuan{90.8\%} \\
\Wenchuan{25} & \Wenchuan{DCRS} & \Wenchuan{0.340} & \Wenchuan{0.145} & \Wenchuan{\textbf{0.278}\textsuperscript{\textbf{3}}} & \Wenchuan{0.043} & \Wenchuan{0.115} & \Wenchuan{0.021} & \Wenchuan{0.178} & \Wenchuan{81.1\%} \\
\addlinespace[1.2pt]
\Wenchuan{26} & \Wenchuan{FINDER} & \Wenchuan{0.153} & \Wenchuan{0.092} & \Wenchuan{0.400} & \Wenchuan{0.051} & \Wenchuan{0.117} & \Wenchuan{0.040} & \Wenchuan{0.135} & \Wenchuan{100.0\%} \\
\Wenchuan{27} & \Wenchuan{MIND} & \Wenchuan{0.147} & \Wenchuan{0.086} & \Wenchuan{0.377} & \Wenchuan{0.049} & \Wenchuan{0.096} & \Wenchuan{0.034} & \Wenchuan{0.118} & \Wenchuan{85.4\%} \\
\addlinespace[1.2pt]
\Wenchuan{28} & \Wenchuan{LLM4CN} & \Wenchuan{0.195} & \Wenchuan{0.103} & \Wenchuan{0.291} & \Wenchuan{0.108} & \Wenchuan{0.074} & \Wenchuan{0.080} & \Wenchuan{0.098} & \Wenchuan{54.6\%} \\
\addlinespace[1.2pt]
\Wenchuan{29} & \Wenchuan{CNT} & \Wenchuan{0.771} & \Wenchuan{0.398} & \Wenchuan{0.351} & \Wenchuan{0.024} & \Wenchuan{0.396} & \Wenchuan{0.008} & \Wenchuan{0.463} & \Wenchuan{77.3\%} \\
\Wenchuan{30} & \Wenchuan{AGNN} & \Wenchuan{0.392} & \Wenchuan{0.192} & \Wenchuan{0.347} & \Wenchuan{0.061} & \Wenchuan{0.252} & \Wenchuan{0.043} & \Wenchuan{0.275} & \Wenchuan{81.1\%} \\
\bottomrule
\end{tabular}%
}
\end{table*}

\paragraph{RQ3: How do different evaluation metrics affect the assessment of network dismantling methods?}\mbox{}\\
\Wenchuan{\textbf{Motivation and Experiment Design.}
Network dismantling performance can be evaluated from different perspectives. Target-based evaluation measures the number of removals required to reach a prescribed fragmentation level, trajectory-based evaluation measures how effectively connectivity is reduced throughout the dismantling process, and structural diagnostics characterize when and how components split. Since these metrics reward different properties of a removal sequence, they may produce different conclusions about method performance. We compare method rankings and rank changes across all three perspectives to determine how strongly the metric choice affects the resulting assessment.}

\paragraph{RQ4: What are the time and space costs of network dismantling methods?}\mbox{}\\
\Wenchuan{\textbf{Motivation and Experiment Design.}
Dismantling quality alone does not determine whether a method can be applied to large real-world networks. Different methods incur substantially different computational costs during training, graph-specific adaptation or search, inference, and post-processing. We therefore compare the observed wall-clock time and peak memory consumption of all methods, while retaining timeout, out-of-memory, and pre-declared skip outcomes as part of the analysis. This evaluation characterizes the trade-off between dismantling effectiveness, computational efficiency, and practical applicability.}

\FloatBarrier
\section{\Wenchuan{Experimental Results and Analysis}}
\label{sec:results}

\subsection{\Wenchuan{Effectiveness of Learning-based Methods (RQ1)}}
\label{sec:rq1}

\paragraph{\Wenchuan{Settings.}}
\Wenchuan{We rank methods on each network and average these failure-aware ranks, placing unsuccessful methods after all successful ones. Table~\ref{tab:main_results_full} reports medians over successful networks together with completion coverage.}

\paragraph{\Wenchuan{Results.}}
\Wenchuan{BPD, CoreHD, and Min-Sum lead the target metric, while Explosive Immunization gives the best LCC-AUC rank. The strongest learning-based methods remain behind these baselines once failed runs are included. FINDER is the only learning-based method that completes all 185 networks; the others trade competitive successful-run results for lower coverage.}

\keyinsights{RQ1}{\Wenchuan{Learning-based methods do not yet outperform strong non-learning methods consistently. Their best results are competitive, but no learning-based method combines leading dismantling performance with complete coverage across the 185 networks.}}

\FloatBarrier

\subsection{\Wenchuan{Variation across Network Domains and Scales (RQ2)}}
\label{sec:rq2}

\begin{figure*}[t]
	\centering
	\includegraphics[width=\textwidth]{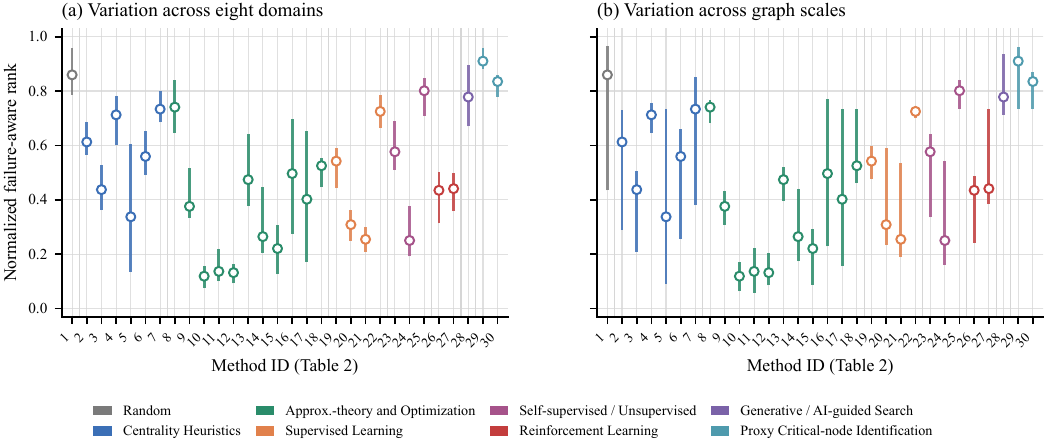}
	\caption{\Wenchuan{Method sensitivity to network domain and scale. Circles denote the overall normalized failure-aware target rank; vertical segments span the best and worst group-specific ranks across eight domains (left) or four size ranges (right). Lower is better. Method IDs refer to Table~\ref{tab:main_results_full}.}}
	\label{fig:domain-scale}
\end{figure*}

\paragraph{\Wenchuan{Settings.}}
\Wenchuan{We compare failure-aware target ranks across eight application domains and four size ranges ($<1$K, 1K--10K, 10K--100K, and $\geq100$K nodes). Figure~\ref{fig:domain-scale} reports method-level variation, while Appendix Table~\ref{tab:category_domain_scale} provides fixed method-family representatives.}

\paragraph{\Wenchuan{Results.}}
\Wenchuan{The strongest methods remain stable across most domains, with BPD leading the fixed family representatives in seven of eight domains. Scale produces larger changes: Adaptive Betweenness performs well below 10K nodes but completes no network above 100K nodes, whereas BPD and CoreHD remain competitive throughout. FINDER covers all large networks at the cost of weaker dismantling quality on smaller ones.}

\keyinsights{RQ2}{\Wenchuan{Scale matters more than domain. Some methods that perform well on small networks become impractical or fail completely as the network grows. Results on large networks should therefore be reported together with completion coverage.}}

\subsection{\Wenchuan{Metric-dependent Assessment (RQ3)}}
\label{sec:rq3}

\paragraph{\Wenchuan{Settings.}}
\Wenchuan{Appendix Table~\ref{tab:metric_top5} compares the five highest-ranked methods under the target criterion and two trajectory summaries, while Appendix Figure~\ref{fig:metric-ranks-appendix} reports all seven rank profiles. Transition diagnostics describe how fragmentation occurs rather than define independent measures of overall superiority.}

\paragraph{\Wenchuan{Results.}}
\Wenchuan{BPD, CoreHD, and Min-Sum rank highly by the target criterion but poorly by PairConn, whereas Static Betweenness, PageRank, and DCRS show the opposite pattern. Explosive Immunization performs well under both target- and trajectory-based evaluation, and the structural diagnostics distinguish gradual fragmentation from a few large component splits.}

\keyinsights{RQ3}{\Wenchuan{Different metrics reward different removal sequences. Target and trajectory measures should be reported together, while structural diagnostics are most useful for explaining why their rankings differ.}}

\subsection{\Wenchuan{Observed Computational Cost and Scalability (RQ4)}}
\label{sec:rq4}

\begin{figure*}[t]
	\centering
	\includegraphics[width=\textwidth]{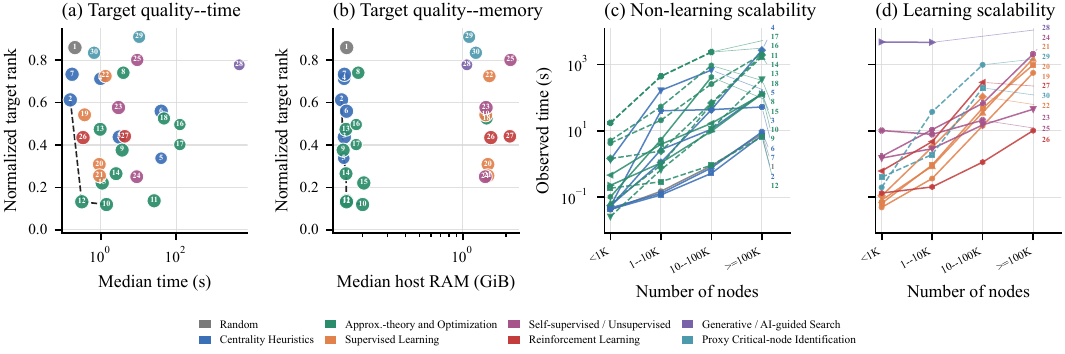}
	\caption{\Wenchuan{Observed target-quality cost trade-offs and runtime scaling over 185 networks. Panels (a) and (b) compare the target rank with median successful-run time and peak host memory; lower and further left is better, dashed lines mark non-dominated observations, and marker area encodes coverage. Panels (c) and (d) show median time by network size for non-learning and learning-based methods, respectively; absent size ranges are not connected. Method IDs refer to Table~\ref{tab:main_results_full}.}}
	\label{fig:cost-scaling}
\end{figure*}

\paragraph{\Wenchuan{Settings.}}
\Wenchuan{Figure~\ref{fig:cost-scaling} pairs failure-aware target rank with median successful-run time and peak host memory, retains coverage in marker area, and reports runtime across four size ranges. Runtime includes per-graph adaptation and search; heterogeneous CPU and GPU implementations support practical within-protocol comparisons rather than hardware-independent complexity claims. Trajectory-based results are provided in Appendix~\ref{app:additional}.}

\paragraph{\Wenchuan{Results.}}
\Wenchuan{CoreHD and BPD combine strong target-based results with median running times below two seconds, while Explosive Immunization achieves the best trajectory rank in about one second. Among learning-based methods, FINDER is fast and complete, whereas methods with better dismantling quality lose coverage on larger networks. LLM4CN marks the opposite extreme, with a median runtime of 4,676 seconds and 54.6\% coverage.}

\keyinsights{RQ4}{\Wenchuan{More computation does not produce better dismantling performance in our experiments. Several non-learning methods are both faster and more effective than the learning-based alternatives. A low median runtime does not necessarily indicate good scalability when a method fails to complete many large networks. Runtime should therefore be reported together with completion coverage.}}

\FloatBarrier

\section{Conclusion and Future Directions}

\Wenchuan{This paper presents \textsc{NetBreaker}, a comprehensive benchmark of 30 network dismantling methods on 185 networks under a common execution and evaluation protocol. \textsc{NetBreaker} converts different method outputs into node removal sequences, enabling direct comparison between non-learning and learning-based methods through the same evaluation procedure. The results show that current learning-based methods can be competitive in selected settings, but do not consistently outperform strong non-learning methods. Method rankings vary across evaluation metrics, while network scale affects both dismantling performance and completion coverage. These findings show that progress in network dismantling cannot be adequately assessed using a single metric or a limited collection of networks.}

\Wenchuan{Our findings leads to some promising directions for future research:}
	
\Wenchuan{\textbf{Multi-objective learning.} Target-based evaluation, trajectory-based evaluation, and structural transitions describe different properties of a removal sequence. Optimizing one property does not necessarily improve the others. Future learning-based methods should therefore explore learning frameworks that coordinate multiple dismantling objectives while preserving the distinctions among final intervention cost, connectivity throughout the removal process, and the structural process of fragmentation.}
	
\Wenchuan{\textbf{Reliable scaling and coverage.} Performance degradation and incomplete execution on large networks remain important limitations of current methods. Future research should develop methods that generalize reliably across network scales and retain high completion coverage as network size increases. Promising directions include training on networks with diverse scales, learning representations invariant to graph size, and designing proper inductive inference procedures.}
	
\Wenchuan{\textbf{Efficiency-aware method development.} Strong non-learning methods continue to provide favorable trade-offs between dismantling effectiveness and running time. Future learning-based methods should pursue clear improvements across effectiveness, completion coverage, running time, and memory consumption, rather than improving one aspect at the expense of the others. The \textsc{NetBreaker} codebase will provide a common basis for measuring these advances.}

\clearpage
\bibliography{references}

@article{braunstein2016network,
  title={Network dismantling},
  author={Braunstein, Alfredo and Dall'Asta, Luca and Semerjian, Guilhem and Zdeborov{\'a}, Lenka},
  journal={Proceedings of the National Academy of Sciences},
  volume={113},
  number={44},
  pages={12368--12373},
  year={2016}
}

@article{artime2024robustness,
  title={Robustness and resilience of complex networks},
  author={Artime, Oriol and Grassia, Marco and De Domenico, Manlio and Gleeson, James P. and Makse, Hern{\'a}n A. and Mangioni, Giuseppe and Perc, Matja{\v z} and Radicchi, Filippo},
  journal={Nature Reviews Physics},
  volume={6},
  number={2},
  pages={114--131},
  year={2024},
  doi={10.1038/s42254-023-00676-y}
}

@article{wandelt2018comparative,
  title={A comparative analysis of approaches to network-dismantling},
  author={Wandelt, Sebastian and Sun, Xiaoqian and Feng, Daozhong and Zanin, Massimiliano and Havlin, Shlomo},
  journal={Scientific Reports},
  volume={8},
  number={1},
  pages={13513},
  year={2018},
  doi={10.1038/s41598-018-31902-8}
}

@inproceedings{kunegis2013konect,
  title={{KONECT}: The Koblenz Network Collection},
  author={Kunegis, J{\'e}r\^ome},
  booktitle={Proceedings of the 22nd International Conference on World Wide Web Companion},
  pages={1343--1350},
  year={2013},
  doi={10.1145/2487788.2488173}
}

@misc{leskovec2014snap,
  title={{SNAP} Datasets: Stanford Large Network Dataset Collection},
  author={Leskovec, Jure and Krevl, Andrej},
  year={2014},
  howpublished={\url{https://snap.stanford.edu/data}}
}

@inproceedings{rossi2015network,
  title={The Network Data Repository with Interactive Graph Analytics and Visualization},
  author={Rossi, Ryan A. and Ahmed, Nesreen K.},
  booktitle={Proceedings of the Twenty-Ninth AAAI Conference on Artificial Intelligence},
  pages={4292--4293},
  year={2015}
}

@article{sen2008collective,
  title={Collective Classification in Network Data},
  author={Sen, Prithviraj and Namata, Galileo and Bilgic, Mustafa and Getoor, Lise and Gallagher, Brian and Eliassi-Rad, Tina},
  journal={AI Magazine},
  volume={29},
  number={3},
  pages={93--106},
  year={2008}
}

@inproceedings{adler2026lgdna,
  title={Latent Geometry-Driven Network Automata for Complex Network Dismantling},
  author={Adler, Thomas and Grassia, Marco and Liao, Ziheng and Mangioni, Giuseppe and Cannistraci, Carlo Vittorio},
  booktitle={International Conference on Learning Representations},
  year={2026}
}

@article{morone2015collective,
  title={Influence maximization in complex networks through optimal percolation},
  author={Morone, Flaviano and Makse, Hern{\'a}n A.},
  journal={Nature},
  volume={524},
  pages={65--68},
  year={2015}
}

@article{zdeborova2016fast,
  title={Fast and simple decycling and dismantling of networks},
  author={Zdeborov{\'a}, Lenka and Zhang, Pan and Zhou, Hai-Jun},
  journal={Scientific Reports},
  volume={6},
  pages={37954},
  year={2016}
}

@article{mugisha2016bpd,
  title={Identifying optimal targets of network attack by belief propagation},
  author={Mugisha, Sheila and Zhou, Hai-Jun},
  journal={Physical Review E},
  volume={94},
  pages={012305},
  year={2016}
}

@article{clusella2016explosive,
  title={Immunization and targeted destruction of networks using explosive percolation},
  author={Clusella, Pau and Grassberger, Peter and P{\'e}rez-Reche, Francisco J. and Politi, Antonio},
  journal={Physical Review Letters},
  volume={117},
  pages={208301},
  year={2016}
}

@article{ren2019gnd,
  title={Generalized network dismantling},
  author={Ren, Xiao-Long and Gleinig, Niels and Helbing, Dirk and Antulov-Fantulin, Nino},
  journal={Proceedings of the National Academy of Sciences},
  volume={116},
  number={14},
  pages={6554--6559},
  year={2019}
}

@article{fan2020finder,
  title={Finding key players in complex networks through deep reinforcement learning},
  author={Fan, Changjun and Zeng, Li and Sun, Yizhou and Liu, Yang-Yu},
  journal={Nature Machine Intelligence},
  volume={2},
  pages={317--324},
  year={2020}
}

@article{grassia2021gdm,
  title={Machine learning dismantling and early-warning signals of disintegration in complex systems},
  author={Grassia, Marco and De Domenico, Manlio and Mangioni, Giuseppe},
  journal={Nature Communications},
  volume={12},
  pages={5190},
  year={2021}
}

@inproceedings{zhang2022nirm,
  title={Dismantling complex networks by a neural model trained from tiny networks},
  author={Zhang, Junhao and Wang, Biao},
  booktitle={Proceedings of the ACM SIGKDD Conference on Knowledge Discovery and Data Mining},
  year={2022}
}

@inproceedings{zhang2023dcrs,
  title={Encoding node diffusion competence and role significance for network dismantling},
  author={Zhang, Junhao and Wang, Biao},
  booktitle={Proceedings of the ACM Web Conference},
  year={2023}
}

@inproceedings{tian2026mind,
  title={Learning network dismantling without handcrafted inputs},
  author={Tian, Hao and Ferraro, Pietro and Shorten, Robert and Jalili, Mahdi and Hamedmoghadam, Homayoun},
  booktitle={Proceedings of the AAAI Conference on Artificial Intelligence},
  year={2026}
}

@article{zhang2026stable,
  title={Neural algorithmic approach to network dismantling},
  author={Zhang, Pan and Fu, Jin-Hua and Sun, Xiao and Cheng, Tao and Chen, Guanrong},
  journal={Science China Information Sciences},
  volume={69},
  pages={119203},
  year={2026}
}

@article{chen2024cnt,
  title={Identifying influential nodes in complex networks via Transformer},
  author={Chen, Liang and others},
  journal={Information Processing \& Management},
  volume={61},
  pages={103775},
  year={2024}
}

@article{xiong2024agnn,
  title={Vital node identification in complex networks based on autoencoder and graph neural network},
  author={Xiong, Y. and Hu, Z. and Su, C. and Cai, S.-M. and Zhou, T.},
  journal={Applied Soft Computing},
  volume={163},
  pages={111895},
  year={2024}
}

@article{holme2002attack,
  title={Attack vulnerability of complex networks},
  author={Holme, Petter and Kim, Beom Jun and Yoon, Chang No and Han, Seung Kee},
  journal={Physical Review E},
  volume={65},
  pages={056109},
  year={2002},
  doi={10.1103/PhysRevE.65.056109}
}

@techreport{page1998pagerank,
  title={The {PageRank} Citation Ranking: Bringing Order to the Web},
  author={Page, Lawrence and Brin, Sergey and Motwani, Rajeev and Winograd, Terry},
  institution={Stanford University},
  year={1998}
}

@article{kitsak2010influential,
  title={Identification of influential spreaders in complex networks},
  author={Kitsak, Maksim and Gallos, Lazaros K. and Havlin, Shlomo and Liljeros, Fredrik and Muchnik, Lev and Stanley, H. Eugene and Makse, Hern{\'a}n A.},
  journal={Nature Physics},
  volume={6},
  pages={888--893},
  year={2010},
  doi={10.1038/nphys1746}
}

@article{fan2021cycle,
  title={Characterizing cycle structure in complex networks},
  author={Fan, Tianlong and L{\"u}, Linyuan and Shi, Dinghua and Zhou, Tao},
  journal={Communications Physics},
  volume={4},
  pages={272},
  year={2021},
  doi={10.1038/s42005-021-00781-3}
}

@article{ren2019ensemble,
  title={Ensemble approach for generalized network dismantling},
  author={Ren, Xiao-Long and Antulov-Fantulin, Nino},
  journal={arXiv preprint arXiv:1909.09016},
  year={2019},
  url={https://arxiv.org/abs/1909.09016}
}

@incollection{coregdm2023,
  title={{CoreGDM}: Geometric Deep Learning Network Decycling and Dismantling},
  author={Grassia, Marco and Mangioni, Giuseppe},
  booktitle={Complex Networks XIV},
  series={Springer Proceedings in Complexity},
  pages={86--94},
  publisher={Springer},
  year={2023},
  doi={10.1007/978-3-031-28276-8_8}
}

@article{mao2024llm4cn,
  title={Identify Critical Nodes in Complex Network with Large Language Models},
  author={Mao, Jinzhu and Zou, Dongyun and Sheng, Li and Liu, Siyi and Gao, Chen and Wang, Yue and Li, Yong},
  journal={arXiv preprint arXiv:2403.03962},
  year={2024},
  url={https://arxiv.org/abs/2403.03962}
}

@article{joyce2010leverage,
  title={A New Measure of Centrality for Brain Networks},
  author={Joyce, Karen E. and Laurienti, Paul J. and Burdette, Jonathan H. and Hayasaka, Satoru},
  journal={PLOS ONE},
  volume={5},
  number={8},
  pages={e12200},
  year={2010},
  doi={10.1371/journal.pone.0012200}
}

@article{chen2013clusterrank,
  title={Identifying Influential Nodes in Large-Scale Directed Networks: The Role of Clustering},
  author={Chen, Duan-Bing and Gao, Hui and L{\"u}, Linyuan and Zhou, Tao},
  journal={PLOS ONE},
  volume={8},
  number={10},
  pages={e77455},
  year={2013},
  doi={10.1371/journal.pone.0077455}
}

@article{xu2020entropy,
  title={Identifying vital nodes in complex networks by adjacency information entropy},
  author={Xu, Xiang and Zhu, Cheng and Wang, Qingyong and Zhu, Xianqiang and Zhou, Yun},
  journal={Scientific Reports},
  volume={10},
  pages={2691},
  year={2020},
  doi={10.1038/s41598-020-59616-w}
}

@article{cohen2003acquaintance,
  title={Efficient Immunization Strategies for Computer Networks and Populations},
  author={Cohen, Reuven and Havlin, Shlomo and ben-Avraham, Daniel},
  journal={Physical Review Letters},
  volume={91},
  pages={247901},
  year={2003},
  doi={10.1103/PhysRevLett.91.247901}
}

@article{osat2023embedding,
  title={Embedding-aided network dismantling},
  author={Osat, Saeed and Papadopoulos, Fragkiskos and Teixeira, Andreia Sofia and Radicchi, Filippo},
  journal={Physical Review Research},
  volume={5},
  pages={013076},
  year={2023},
  doi={10.1103/PhysRevResearch.5.013076}
}

@article{wu2025mhrl,
  title={A reinforcement learning-enhanced meta-heuristic framework for network dismantling},
  author={Wu, Min and Shi, Wu and Guo, Fengwei and Dai, Bitao and Mou, Jianhong and Tan, Suoyi and Lu, Xin and Ou, Chaomin},
  journal={Journal of Physics: Complexity},
  volume={6},
  pages={035012},
  year={2025},
  doi={10.1088/2632-072X/adfff1}
}

@article{deng2025gagcn,
  title={A hybrid approach to network disintegration: Integrating graph convolutional network and genetic algorithm},
  author={Deng, Ye and Tan, Dingrong and Shen, Xiaoda and Wang, Zhigang and Wu, Jun},
  journal={Chaos, Solitons \& Fractals},
  volume={197},
  pages={116488},
  year={2025},
  doi={10.1016/j.chaos.2025.116488}
}

@article{han2025glnd,
  title={Graph Learning-Based Network Dismantling via hierarchical feature fusion and margin-ranked optimization},
  author={Han, Jihui and Shi, Yupeng and Huang, Daoying and Huang, Ziqi and Zhang, Anlin},
  journal={Journal of Complex Networks},
  volume={13},
  number={4},
  pages={cnaf015},
  year={2025},
  doi={10.1093/comnet/cnaf015}
}

@article{liu2022nees,
  title={Neural extraction of multiscale essential structure for network dismantling},
  author={Liu, Qingxia and Wang, Bang},
  journal={Neural Networks},
  volume={154},
  pages={99--108},
  year={2022},
  doi={10.1016/j.neunet.2022.07.015}
}

@article{hong2025gasc,
  title={A deep learning dismantling approach for understanding the structural vulnerability of complex networks},
  author={Hong, Li and Xu, Mengqiao and Liu, Yu and Zhang, Xuemei and Fan, Changjun},
  journal={Chaos, Solitons \& Fractals},
  volume={194},
  pages={116148},
  year={2025},
  doi={10.1016/j.chaos.2025.116148}
}

@article{ma2024clma,
  title={Dismantling complex networks with graph contrastive learning and multi-hop aggregation},
  author={Ma, Siqi and Zeng, Weixin and Xiao, Weidong and Zhao, Xiang},
  journal={Information Sciences},
  volume={676},
  pages={120780},
  year={2024},
  doi={10.1016/j.ins.2024.120780}
}

@article{shi2026cpds,
  title={Complex network dismantling with node diffusion and node similarity},
  author={Shi, Jinchen and Tang, Jiuyang and Tan, Zhen and Hu, Yanli and Fang, Yang},
  journal={Expert Systems with Applications},
  volume={300},
  pages={130433},
  year={2026},
  doi={10.1016/j.eswa.2025.130433}
}

@article{zhou2026spr,
  title={Dismantling complex networks based on higher-order graph neural network},
  author={Zhou, Wennan and Tan, Suoyi and Fang, Yang and L{\"u}, Xin and Zhao, Xiang},
  journal={Communications Physics},
  volume={9},
  pages={181},
  year={2026},
  doi={10.1038/s42005-026-02601-y}
}

@article{fan2020coregqn,
  title={A novel learning-based approach for efficient dismantling of networks},
  author={Fan, Changjun and Zeng, Li and Feng, Yanghe and Cheng, Guangquan and Huang, Jincai and Liu, Zhong},
  journal={International Journal of Machine Learning and Cybernetics},
  volume={11},
  pages={2101--2111},
  year={2020},
  doi={10.1007/s13042-020-01104-8}
}

@article{pu2024smartcore,
  title={Deep Reinforcement Learning for Network Dismantling: A {K}-Core Based Approach},
  author={Pu, Tianle and Zeng, Li and Chen, Chao},
  journal={Mathematics},
  volume={12},
  pages={1215},
  year={2024},
  doi={10.3390/math12081215}
}

@article{xing2023peerremove,
  title={{PeerRemove}: An adaptive node removal strategy for {P2P} botnet based on deep reinforcement learning},
  author={Xing, Ying and Shu, Hui and Kang, Fei},
  journal={Computers \& Security},
  volume={128},
  pages={103129},
  year={2023},
  doi={10.1016/j.cose.2023.103129}
}

@inproceedings{qu2025grlnd,
  title={{GRLND}: A Graph Reinforcement Learning Framework for Network Dismantling},
  author={Qu, Hongbo and Wang, Xu and Song, Yu-Rong and Ni, Wei and Jiang, Guo-Ping and Sheng, Quan Z.},
  booktitle={Proceedings of the 34th ACM International Conference on Information and Knowledge Management},
  pages={2450--2459},
  year={2025},
  doi={10.1145/3746252.3761371}
}

@article{fu2026gfd,
  title={Exploring diverse solutions in network dismantling with generative model},
  author={Fu, Jun and Sun, Xiaojie and Zhang, Peng and Findeisen, Rolf},
  journal={Chaos, Solitons \& Fractals},
  volume={202},
  pages={117526},
  year={2026},
  doi={10.1016/j.chaos.2025.117526}
}

@article{chen2023deepele,
  title={Identifying critical nodes via link equations and deep reinforcement learning},
  author={Chen, Peiyu and Fan, Wenhui},
  journal={Neurocomputing},
  volume={562},
  pages={126871},
  year={2023},
  doi={10.1016/j.neucom.2023.126871}
}

@article{zhao2023honnma,
  title={A novel higher-order neural network framework based on motifs attention for identifying critical nodes},
  author={Zhao, Xiuming and Yu, Hongtao and Huang, Ruiyang and Liu, Shuxin and Hu, Nan and Cao, Xiaochun},
  journal={Physica A: Statistical Mechanics and its Applications},
  volume={629},
  pages={129194},
  year={2023},
  doi={10.1016/j.physa.2023.129194}
}

@article{munikoti2022ilgr,
  title={Scalable graph neural network-based framework for identifying critical nodes and links in complex networks},
  author={Munikoti, Sai and Das, Laya and Natarajan, Balasubramaniam},
  journal={Neurocomputing},
  volume={468},
  pages={211--221},
  year={2022},
  doi={10.1016/j.neucom.2021.10.031}
}

@article{huang2026stgcn,
  title={A Sparse Transformer-Enhanced Graph Convolutional Model for Robust Node Importance Ranking in Complex Networks},
  author={Huang, Ru and Zhou, Shuo and Li, Pengfei and Yang, Kun and Chen, Zijian and He, Jianhua and Chu, Xiaoli and Zhou, Zhengbing and Zhai, Guangtao},
  journal={IEEE Transactions on Network Science and Engineering},
  volume={13},
  pages={2350--2370},
  year={2026},
  doi={10.1109/TNSE.2025.3614439}
}
\bibliographystyle{iclr2027_conference}

\clearpage
\appendix
\clearpage
\section{Benchmark Dataset Details}
\label{app:datasets}

\subsection{\Wenchuan{The Taxonomy of Dataset Domains}}

\Wenchuan{Existing network dismantling studies use substantially inconsistent taxonomies for dataset domains. Some studies provide no explicit categorization, whereas others organize networks using either a few broad domains or multi-level schemes with dozens of types; even similarly named categories often differ in scope. Retaining these study-specific labels would therefore make domain coverage incomparable across studies. We consequently reclassify every real-world dataset according to the system represented by the network, rather than its source repository, file format, or edge type. Table~\ref{tab:prior-domain-taxonomies} summarizes the schemes adopted in existing studies, and Table~\ref{tab:unified-domain-taxonomy} defines the eight-domain hierarchy used in our benchmark. Each real-world network is assigned one primary domain and one sub-domain, while the synthetic reference is recorded separately and excluded from domain coverage.}

\begin{table}[H]
\caption{\Wenchuan{The Taxonomy of Dataset Domains Adopted in Existing Network Dismantling Studies.}}
\label{tab:prior-domain-taxonomies}
\centering
\footnotesize
\setlength{\tabcolsep}{4pt}
\renewcommand{\arraystretch}{1.08}
\begin{tabular}{@{}>{\raggedright\arraybackslash}p{0.23\linewidth}>{\raggedright\arraybackslash}p{0.22\linewidth}>{\raggedright\arraybackslash}p{0.49\linewidth}@{}}
\toprule
\Wenchuan{Study} & \Wenchuan{Categorization Scheme} & \Wenchuan{Domain Categories} \\
\midrule
\Wenchuan{Wandelt et al.\ \citep{wandelt2018comparative}} & \Wenchuan{Not provided} & \Wenchuan{---} \\
\Wenchuan{Artime et al.\ \citep{artime2024robustness}} & \Wenchuan{Single-level, 4 categories} & \Wenchuan{Biological; Information; Social; Technological} \\
\Wenchuan{FINDER \citep{fan2020finder}} & \Wenchuan{Single-level, 5 categories} & \Wenchuan{Criminal; Biological; Communication; Infrastructure; Social} \\
\Wenchuan{GDM \citep{grassia2021gdm}} & \Wenchuan{Single-level, 16 categories} & \Wenchuan{Infrastructure; Social; Metabolic; Hyperlink; Citation; Coauthorship; Neural; Economic; Communication; Power; Trophic; Software; Transport; Human Contact; Computer; Lexical} \\
\Wenchuan{MIND \citep{tian2026mind}} & \Wenchuan{Single-level, 4 categories} & \Wenchuan{Biological; Social; Information; Technological} \\
\Wenchuan{LGD-NA \citep{adler2026lgdna}} & \Wenchuan{Two-level, 8 fields and 32 types} & \Wenchuan{Biomolecular; Brain; Covert; Food Web; Infrastructure; Internet; Miscellaneous; Social} \\
\bottomrule
\end{tabular}
\end{table}

\begin{table}[H]
\caption{\Wenchuan{The unified taxonomy of dataset domains and sub-domains used in this study.}}
\label{tab:unified-domain-taxonomy}
\centering
\footnotesize
\setlength{\tabcolsep}{5pt}
\renewcommand{\arraystretch}{1.12}
\begin{tabular}{@{}>{\raggedright\arraybackslash}p{0.35\linewidth}>{\raggedright\arraybackslash}p{0.60\linewidth}@{}}
\toprule
\Wenchuan{Domains} & \Wenchuan{Sub-domains} \\
\midrule
\Wenchuan{Biological and Ecological} & \Wenchuan{Protein and molecular interaction; Gene regulation; Metabolic interaction; Neural connectivity; Drug interaction; Food webs; Animal social interaction} \\
\Wenchuan{Social and Human Interaction} & \Wenchuan{Friendship and online social interaction; Trust and reputation; Human communication; Physical contact; Professional and organizational interaction; Location-based interaction; Political and geopolitical interaction; Entertainment and cultural interaction; Sports interaction} \\
\Wenchuan{Information, Knowledge, and Media} & \Wenchuan{Web and hyperlinks; Online media; Knowledge and encyclopedia; Lexical and linguistic; Literary and fictional; Collaborative editing and voting} \\
\Wenchuan{Scientific and Scholarly} & \Wenchuan{Citation; Coauthorship; Scientific collaboration} \\
\Wenchuan{Technological and Communication Infrastructure} & \Wenchuan{Internet and autonomous systems; Peer-to-peer systems; Power grids; Software dependency; Electronic circuits} \\
\Wenchuan{Transportation and Mobility} & \Wenchuan{Roads; Airlines; Urban transit; Maritime shipping; Traffic and geographic mobility} \\
\Wenchuan{Economic and Financial} & \Wenchuan{Trade; Commerce and co-purchasing; Financial trust and transactions} \\
\Wenchuan{Security, Crime, and Covert} & \Wenchuan{Crime; Terrorism; Corruption; Botnets and malicious activity} \\
\bottomrule
\end{tabular}
\end{table}

\subsection{Network Collection, Preprocessing, and Statistics}

\Wenchuan{The benchmark contains 185 distinct processed graphs: 184 real-world networks organized under the eight-domain taxonomy in Table~\ref{tab:unified-domain-taxonomy}, together with one synthetic reference. Table~\ref{tab:dataset-summary} summarizes the number, processed scale, median graph-average degree, and size distribution of the networks in each domain. The accompanying dataset registry records per-network identifiers, provenance, domain and sub-domain assignments, processed node and edge counts, preprocessing outcomes, and structural statistics.}

\begin{table}[H]
\caption{\Wenchuan{Dataset statistics for 184 real-world networks across eight unified domains and one synthetic reference. Node and edge counts refer to processed simple, undirected largest connected components. The four size counts correspond to $N<10^3$, $10^3\leq N<10^4$, $10^4\leq N<10^5$, and $N\geq10^5$, matching the scale analysis in Section~\ref{sec:rq2}. The degree column reports the median graph-average degree; k and M denote $10^3$ and $10^6$.}}
\label{tab:dataset-summary}
\centering
\scriptsize
\setlength{\tabcolsep}{3pt}
\renewcommand{\arraystretch}{1.08}
\begin{tabular}{@{}>{\raggedright\arraybackslash}p{0.31\linewidth}rrrrr@{}}
\toprule
\Wenchuan{Domain} & \Wenchuan{Graphs} & \Wenchuan{Nodes: range} & \Wenchuan{Edges: range} & \Wenchuan{Med. $\langle k\rangle$} & \Wenchuan{Size counts} \\
\midrule
\Wenchuan{Biological and Ecological} & \Wenchuan{26} & \Wenchuan{14--324.2k} & \Wenchuan{20--2.64M} & \Wenchuan{7.74} & \Wenchuan{11/13/1/1} \\
\Wenchuan{Social and Human Interaction} & \Wenchuan{54} & \Wenchuan{49--1.63M} & \Wenchuan{107--22.30M} & \Wenchuan{10.39} & \Wenchuan{15/12/14/13} \\
\Wenchuan{Information, Knowledge, and Media} & \Wenchuan{22} & \Wenchuan{77--855.8k} & \Wenchuan{167--4.29M} & \Wenchuan{7.52} & \Wenchuan{5/9/4/4} \\
\Wenchuan{Scientific and Scholarly} & \Wenchuan{23} & \Wenchuan{86--365.2k} & \Wenchuan{124--1.72M} & \Wenchuan{7.94} & \Wenchuan{2/6/13/2} \\
\Wenchuan{Technological and Communication Infrastructure} & \Wenchuan{28} & \Wenchuan{512--1.69M} & \Wenchuan{819--11.09M} & \Wenchuan{5.25} & \Wenchuan{1/13/12/2} \\
\Wenchuan{Transportation and Mobility} & \Wenchuan{18} & \Wenchuan{39--1.35M} & \Wenchuan{129--1.88M} & \Wenchuan{3.42} & \Wenchuan{9/6/1/2} \\
\Wenchuan{Economic and Financial} & \Wenchuan{6} & \Wenchuan{105--334.9k} & \Wenchuan{441--925.9k} & \Wenchuan{7.94} & \Wenchuan{3/2/0/1} \\
\Wenchuan{Security, Crime, and Covert} & \Wenchuan{7} & \Wenchuan{64--4,801} & \Wenchuan{243--882.6k} & \Wenchuan{7.59} & \Wenchuan{5/2/0/0} \\
\midrule
\Wenchuan{Synthetic reference} & \Wenchuan{1} & \Wenchuan{84} & \Wenchuan{775} & \Wenchuan{18.45} & \Wenchuan{1/0/0/0} \\
\midrule
\Wenchuan{All networks} & \Wenchuan{185} & \Wenchuan{14--1.69M} & \Wenchuan{20--22.30M} & \Wenchuan{7.35} & \Wenchuan{52/63/45/25} \\
\bottomrule
\end{tabular}
\end{table}

\Wenchuan{All inputs are converted into simple, undirected, and unweighted graphs: self-loops are removed, duplicate or reciprocal edge records are merged, the largest connected component is retained, and node identifiers are relabeled consistently. Source-level and processed counts are both preserved to document these transformations. Networks whose meaning depends on edge direction, weights, temporal order, or multiple layers are excluded rather than treated as equivalent static inputs. Among 219 catalogue records, 213 were processed successfully; collapsing 28 provenance aliases yields 185 distinct graphs, while four failed records and two quarantined records are excluded.}

\Wenchuan{For each network, we retain the unified domain label and processed node count used in Sections~\ref{sec:rq2} and~\ref{sec:rq4}, together with edge count, average degree, degree coefficient of variation, average clustering, and degree assortativity for reproducibility and structural diagnostics. The synthetic graph is labeled explicitly as a reference rather than presented as a systematic synthetic benchmark. Dataset-level provenance and any reported license information follow the original documentation and are not inferred from file availability.}

\section{Benchmark Algorithms}
\label{app:algorithms}

\subsection{\Wenchuan{Method Taxonomy and Inventory}}

\Wenchuan{This appendix first summarizes the taxonomy and implementation provenance of the 30 evaluated methods, and then introduces each method according to the same family organization used throughout the benchmark. Methods reviewed during benchmark construction but not included in the evaluation are listed separately at the end of the appendix.}

% Tianci (round 3): family table moved here from Section 3.2.
\begin{table}[htbp]
\caption{\Tianci{The 30 evaluated methods by family. IDs refer to Table~\ref{tab:main_results_full}; ``+R'' denotes a reinsertion variant. Full descriptions and implementation sources are in Appendix~\ref{app:algorithms}.}}
\label{tab:method-families}
\centering
\scriptsize
\setlength{\tabcolsep}{3pt}
\renewcommand{\arraystretch}{1.08}
{
\begin{tabular}{@{}>{\raggedright\arraybackslash}p{0.155\linewidth}>{\raggedright\arraybackslash}p{0.36\linewidth}c>{\raggedright\arraybackslash}p{0.22\linewidth}>{\raggedright\arraybackslash}p{0.17\linewidth}@{}}
\toprule
Family & Methods (ID) & $n$ & Removal order formed by & Learned from \\
\midrule
Random control & Random (1) & 1 & seeded permutation & -- \\
Centrality heuristics & Degree and Betweenness (static, adaptive), PageRank, $k$-core (2--7) & 6 & one-pass score, or recomputation after each removal & -- \\
Approx.\ theory and optimization & Cycle Ratio, CI, BPD, Min-Sum, CoreHD, GND(+R), Explosive Imm., Ensemble-GND(+R), LGD-NA CND (8--18) & 11 & cycle analysis, message passing, decycling, or partition search & -- \\
Supervised learning & GDM(+R), CoreGDM, NIRM (19--22) & 4 & GNN score; +R adds greedy reinsertion & small graphs with optimal labels \\
Self-/unsupervised & Stable-AGDM(+R), DCRS (23--25) & 3 & graph scoring or per-graph adaptation & unlabeled graph objectives \\
Reinforcement learning & FINDER, MIND (26--27) & 2 & sequential policy on the residual graph & synthetic graphs \\
Generative search & LLM4CN (28) & 1 & LLM-evolved scoring program per graph & target graph \\
Proxy critical-node & CNT, AGNN (29--30) & 2 & one-pass influence ranking & SIR spreading labels \\
\bottomrule
\end{tabular}}
\end{table}

\Wenchuan{Table~\ref{tab:method-families} provides the family-level taxonomy, while Tables~\ref{tab:methods-nonlearning} and~\ref{tab:methods-learning} enumerate the individual methods and variants. Each ``link'' points to the implementation used or reviewed for the corresponding method.} \Tianci{Availability and licensing were checked at the time of the literature review and will be re-verified before release; a locally retained copy does not by itself establish complete provenance or an explicit license.} % Tianci: the previous sentence named an internal review date (2026-08-23), which does not belong in the paper.

\begin{table}[H]
\caption{Evaluated non-learning methods and variants (18 entries). Modes are one-pass (static), adaptive, or global search.}
\label{tab:methods-nonlearning}
\centering
\scriptsize
\setlength{\tabcolsep}{3.0pt}
\begin{tabular}{@{}rlllll@{}}
\toprule
ID & Method / variant & Subcategory & Mode & Output & Implementation source \\
\midrule
1 & Random & negative control & one-pass & sequence & \href{https://github.com/b010001y/NetBreaker}{link} \\
2 & Static Degree & centrality & one-pass & score & \href{https://github.com/b010001y/NetBreaker}{link} \\
3 & Adaptive Degree & centrality & adaptive & sequence & \href{https://github.com/b010001y/NetBreaker}{link} \\
4 & Static Betweenness & centrality & one-pass & score & \href{https://github.com/networkx/networkx}{link} \\
5 & Adaptive Betweenness & centrality & adaptive & sequence & \href{https://github.com/networkx/networkx}{link} \\
6 & PageRank & centrality & one-pass & score & \href{https://github.com/networkx/networkx}{link} \\
7 & k-core/k-shell & centrality & one-pass & score & \href{https://github.com/networkx/networkx}{link} \\
8 & Cycle Ratio & centrality & one-pass & score & \href{https://github.com/ftl129/CycleRatio}{link} \\
9 & Collective Influence & approximate theory & adaptive & sequence & \href{https://github.com/makselab/Collective-Influence}{link} \\
10 & BPD & approximate theory & adaptive & sequence & \href{https://github.com/HaiJunZhou/dismantling}{link} \\
11 & Min-Sum & optimization & global search & \Wenchuan{set} & \href{https://github.com/abraunst/decycler}{link} \\
12 & CoreHD & approximate theory & adaptive & sequence & \Wenchuan{\href{https://github.com/NetworkDismantling/review/tree/main/network_dismantling/CoreHD}{link}} \\
13 & GND & optimization & global search & set & \href{https://github.com/renxiaolong/Generalized-Network-Dismantling}{link} \\
14 & GNDR & optimization + reinsertion & global search & sequence & \href{https://github.com/renxiaolong/Generalized-Network-Dismantling}{link} \\
15 & Explosive Immunization & approximate theory & adaptive & sequence & \href{https://github.com/pclus/explosive-immunization}{link} \\
16 & Ensemble-GND & optimization ensemble & global search & set & \href{https://github.com/renxiaolong/2019-Ensemble-approach-for-generalized-network-dismantling}{link} \\
17 & Ensemble-GNDR & ensemble + reinsertion & global search & sequence & \href{https://github.com/renxiaolong/2019-Ensemble-approach-for-generalized-network-dismantling}{link} \\
18 & LGD-NA CND & latent-geometry optimization & adaptive & sequence & \href{https://github.com/b010001y/NetBreaker}{link} \\
\bottomrule
\end{tabular}
\end{table}

\begin{table}[H]
\caption{Evaluated learning-based and AI-guided methods and variants (12 entries). CoreGDM uses a third-party implementation, and the available LLM4CN artifact has unresolved provenance and completeness.}
\label{tab:methods-learning}
\centering
\scriptsize
\setlength{\tabcolsep}{2.7pt}
\begin{tabular}{@{}rllllll@{}}
\toprule
ID & Method / variant & Role & Training signal & Mode & Output & Reviewed availability \\
\midrule
19 & GDM & direct & supervised & one-pass & score & \href{https://github.com/NetworkScienceLab/GDM}{link} \\
20 & GDM+R & direct & supervised & one-pass + reinsertion & sequence & \href{https://github.com/NetworkScienceLab/GDM}{link} \\
21 & CoreGDM & direct hybrid & supervised & adaptive & sequence & \href{https://github.com/NetworkDismantling/review}{link} \\
22 & NIRM & direct & supervised & one-pass & score & \href{https://github.com/JiazhengZhang/NIRM}{link} \\
23 & Stable-AGDM & direct & \Wenchuan{supervised} & one-pass & score & \href{https://gitee.com/yzhangpeng/stable-agdm}{link} \\
24 & Stable-AGDM+R & direct & \Wenchuan{supervised} & one-pass + reinsertion & sequence & \href{https://gitee.com/yzhangpeng/stable-agdm}{link} \\
25 & DCRS & direct & self/unsupervised & \Wenchuan{per-graph adaptation + one-pass} & score & \href{https://github.com/JiazhengZhang/DCRS}{link} \\
26 & FINDER & direct & reinforcement & adaptive policy & sequence & \Wenchuan{\href{https://github.com/fanchangjun/FINDER}{link}} \\
27 & MIND & direct & reinforcement & adaptive policy & sequence & \href{https://github.com/HaozheTian/MIND-ND}{link} \\
28 & LLM4CN & direct & generative search & program search & score & \href{https://anonymous.4open.science/r/LLM4CN-6520}{link} \\
29 & CNT & proxy & supervised SIR labels & one-pass & score & \href{https://github.com/cly1022/socialmedia-CNT}{link} \\
30 & AGNN & proxy & \Wenchuan{supervised SIR ranking} & \Wenchuan{per-graph adaptation + one-pass} & score & \href{https://github.com/hbjix/AGNN}{link} \\
\bottomrule
\end{tabular}
\end{table}

\subsection{\Wenchuan{Method Descriptions}}

\subsubsection{\Wenchuan{Non-learning Methods}}

\Wenchuan{The non-learning methods are organized into a random control, centrality heuristics, and approximate-theory or optimization-based approaches.}

\paragraph{\Wenchuan{Random control and centrality heuristics.}}

\begin{itemize}[leftmargin=1.35em,itemsep=0.55em,topsep=0.35em]
\item \textbf{Random} is included as a negative-control dismantling strategy. It generates a seeded random permutation of all nodes and therefore uses neither structural information nor residual-graph adaptation.

\item \textbf{Static Degree} \Wenchuan{\citep{holme2002attack}} ranks nodes once by their degree in the original network. It represents the simplest local-centrality strategy and does not update scores after removals.

\item \textbf{Adaptive Degree} \Wenchuan{\citep{holme2002attack}} repeatedly removes a highest-degree node from the current residual network. By recomputing degrees after each intervention, it captures structural changes that the static ranking ignores.

\item \textbf{Static Betweenness} \Wenchuan{\citep{holme2002attack}} ranks nodes once according to shortest-path betweenness in the original network. It targets nodes that mediate many shortest paths, but does not revise their importance after the topology changes.

\item \textbf{Adaptive Betweenness} \Wenchuan{\citep{holme2002attack}} recomputes shortest-path betweenness on the residual network after each removal. It therefore follows the migration of shortest-path bottlenecks during dismantling at a substantially higher computational cost than the static variant.

\item \textbf{PageRank} \Wenchuan{\citep{page1998pagerank}} assigns node importance through the stationary distribution of a random walk with teleportation. The resulting scores are computed once and converted into a descending removal order.

\item \textbf{k-core/k-shell} \Wenchuan{\citep{kitsak2010influential}} decomposes a network into nested shells and prioritizes nodes with larger core numbers. It captures membership in densely connected regions while producing a one-pass ranking.

\item \textbf{Cycle Ratio} \Wenchuan{\citep{fan2021cycle}} measures node importance from participation in elementary cycles and the overlap among those cycles. It prioritizes nodes whose removal disrupts cyclic connectivity and produces a one-pass ranking.

\end{itemize}

\paragraph{\Wenchuan{Approximate-theory and optimization-based methods.}}

\begin{itemize}[leftmargin=1.35em,itemsep=0.55em,topsep=0.35em]

\item \textbf{Collective Influence} \citep{morone2015collective} is proposed to approximate optimal percolation through a finite-radius score that combines a node's excess degree with the degrees on its boundary. Scores are updated on the residual network to construct an adaptive dismantling sequence.

\item \textbf{BPD} \citep{mugisha2016bpd} is proposed to identify a small feedback vertex set using belief-propagation marginals and decimation. Tree breaking and reinsertion then convert the decycling solution into a dismantling sequence.

\item \textbf{Min-Sum} \citep{braunstein2016network} formulates network decycling as a message-passing optimization problem. It first identifies a decycling set and then breaks the remaining forest and reinserts redundant removals to obtain an ordered dismantling solution.

\item \textbf{CoreHD} \citep{zdeborova2016fast} is proposed as a fast decycling heuristic that repeatedly removes a highest-degree node from the current 2-core. The resulting forest is subsequently broken and processed by reinsertion to form the final dismantling sequence.

\item \textbf{GND} \citep{ren2019gnd} recursively partitions a network through weighted spectral bisection and approximates a vertex cover over the cut edges. The union of selected separator nodes forms its global dismantling set.

\item \textbf{GNDR} \citep{ren2019gnd} extends GND with a reinsertion stage that removes redundant nodes and orders the retained solution. This refinement converts the global candidate set into a sequential dismantling strategy.

\item \textbf{Explosive Immunization} \citep{clusella2016explosive} constructs a removal order by reversing an explosive-percolation growth process. Its score is designed to delay the emergence of a giant connected component and thereby identify nodes important for global connectivity.

\item \textbf{Ensemble-GND} \Wenchuan{\citep{ren2019ensemble}} aggregates multiple randomized GND solutions to reduce dependence on a single spectral partition. Candidate dismantling sets are combined into an ensemble solution without the final reinsertion refinement.

\item \textbf{Ensemble-GNDR} \Wenchuan{\citep{ren2019ensemble}} applies reinsertion to the ensemble GND solution, eliminating redundant removals and producing an ordered sequence. It is evaluated separately because the reinsertion stage changes both solution size and trajectory.

\item \textbf{LGD-NA CND} \citep{adler2026lgdna} uses the CND variant of the latent-geometry-driven neural automaton to construct an adaptive dismantling sequence. We evaluate the released procedure as a complete optimization method through the benchmark's common removal-sequence interface.
\end{itemize}

\subsubsection{\Wenchuan{Learning-based and AI-guided Methods}}

\Wenchuan{The learning-based methods are grouped by their training or search mechanism: supervised learning, self-supervised or unsupervised learning, reinforcement learning, generative search, and proxy critical-node identification.}

\paragraph{\Wenchuan{Supervised learning.}}

\begin{itemize}[leftmargin=1.35em,itemsep=0.55em,topsep=0.35em]
\item \textbf{GDM} \citep{grassia2021gdm} is proposed to learn transferable dismantling scores from high-quality solutions on small training graphs. It uses graph neural representations to rank nodes in an unseen network with a single forward scoring stage.

\item \textbf{GDM+R} \citep{grassia2021gdm} augments the raw GDM ranking with greedy reinsertion. The refinement removes redundant selected nodes and produces a sequence whose terminal set differs from that of the direct score order.

\item \textbf{CoreGDM} \Wenchuan{\citep{coregdm2023}} combines the supervised GDM scorer with a CoreHD-style decycling pipeline. It repeatedly applies learned scores within the current 2-core, followed by tree breaking and reinsertion.

\item \textbf{NIRM} \citep{zhang2022nirm} is proposed to learn transferable node-importance scores from dismantling targets generated on small graphs. It propagates target labels and combines local and global encoders to represent both neighborhood structure and network-level context.

\item \textbf{Stable-AGDM} \citep{zhang2026stable} is proposed to improve the stability of learned dismantling scores across different network topologies. It combines supervised message passing with feature selection and outputs a one-pass node ranking.

\item \textbf{Stable-AGDM+R} \citep{zhang2026stable} applies the released refinement and reinsertion procedure to the Stable-AGDM ranking. The post-processing stage removes redundant interventions and changes the resulting dismantling trajectory.

\end{itemize}

\paragraph{\Wenchuan{Self-supervised and unsupervised learning.}}

\begin{itemize}[leftmargin=1.35em,itemsep=0.55em,topsep=0.35em]

\item \textbf{DCRS} \citep{zhang2023dcrs} is proposed to learn critical-node scores without dismantling labels. It encodes diffusion competence and structural-role significance and fuses them through unlabeled, graph-specific adaptation.

\end{itemize}

\paragraph{\Wenchuan{Reinforcement learning.}}

\begin{itemize}[leftmargin=1.35em,itemsep=0.55em,topsep=0.35em]

\item \textbf{FINDER} \citep{fan2020finder} is proposed to learn an adaptive dismantling policy through deep reinforcement learning. It combines structure2vec representations with an $n$-step Q-function trained on synthetic graphs and applies the learned policy sequentially to unseen networks.

\item \textbf{MIND} \citep{tian2026mind} is proposed to optimize sequential network dismantling through a discrete soft actor--critic framework. It couples graph message passing with an adaptive policy trained against accumulated LCC behavior.

\end{itemize}

\paragraph{\Wenchuan{Generative search and proxy critical-node identification.}}

\begin{itemize}[leftmargin=1.35em,itemsep=0.55em,topsep=0.35em]

\item \textbf{LLM4CN} \Wenchuan{\citep{mao2024llm4cn}} is proposed to search for interpretable node-scoring programs through language-model-guided crossover and mutation. Candidate programs are evaluated by dismantling fitness, and the selected program is then applied to generate node scores. Its provenance and snapshot completeness remain pending verification.

\item \textbf{CNT} \citep{chen2024cnt} is proposed to identify influential nodes by learning from SIR-based supervision. It uses Transformer self-attention to model global node interactions and outputs a one-pass influence ranking, making it a proxy critical-node method rather than a direct LCC-dismantling optimizer.

\item \textbf{AGNN} \citep{xiong2024agnn} is proposed to identify vital nodes by combining an autoencoder, graph message passing, and ranking supervision derived from SIR simulations. It produces a one-pass influence ranking and is therefore evaluated as a proxy critical-node method. The reviewed snapshot has incomplete data and checkpoint dependencies.
\end{itemize}

\subsection{\Wenchuan{Methods Reviewed but Not Evaluated}}

\Wenchuan{Table~\ref{tab:survey-only} records additional methods considered during benchmark construction. These methods are kept separate from the evaluated inventory so that literature coverage is not conflated with executable benchmark coverage.}

\begin{table}[H]
\caption{Additional methods covered by the literature survey but not evaluated in this study. \Wenchuan{``No verified executable implementation'' means that the recorded review did not identify a runnable implementation of the proposed method; it is not a claim that no implementation exists.}}
\label{tab:survey-only}
\centering
\scriptsize
\begin{tabular}{@{}p{.34\textwidth}p{.18\textwidth}p{.08\textwidth}p{.31\textwidth}@{}}
\toprule
Methods & Role/category & Evaluated? & \Wenchuan{Reason not evaluated} \\
\midrule
Leverage \Wenchuan{\citep{joyce2010leverage}}, ClusterRank \Wenchuan{\citep{chen2013clusterrank}}, adjacency entropy \Wenchuan{\citep{xu2020entropy}}, acquaintance \Wenchuan{\citep{cohen2003acquaintance}} & non-learning heuristics & No & \Wenchuan{outside the executable suite} \\
Embedding-aided dismantling \Wenchuan{\citep{osat2023embedding}}, AMH \Wenchuan{\citep{wu2025mhrl}} & non-learning optimization & No & \Wenchuan{not integrated into the executable suite} \\
GA--GCN \Wenchuan{\citep{deng2025gagcn}}, GLND \Wenchuan{\citep{han2025glnd}} & supervised direct & No & \Wenchuan{no verified executable implementation} \\
NEES \Wenchuan{\citep{liu2022nees}}, GASC \Wenchuan{\citep{hong2025gasc}}, CLMA \Wenchuan{\citep{ma2024clma}}, CPDS \Wenchuan{\citep{shi2026cpds}}, SPR \Wenchuan{\citep{zhou2026spr}} & self/unsupervised direct & No & \Wenchuan{no verified executable implementation} \\
CoreGQN \Wenchuan{\citep{fan2020coregqn}}, SmartCore \Wenchuan{\citep{pu2024smartcore}}, PeerRemove \Wenchuan{\citep{xing2023peerremove}}, GRLND \Wenchuan{\citep{qu2025grlnd}}, MHRL \Wenchuan{\citep{wu2025mhrl}} & reinforcement direct & No & \Wenchuan{no verified executable implementation} \\
GFD \Wenchuan{\citep{fu2026gfd}} & generative direct & No & \Wenchuan{no verified executable implementation} \\
DeepELE \Wenchuan{\citep{chen2023deepele}}, HONNMA \Wenchuan{\citep{zhao2023honnma}}, STGCN \Wenchuan{\citep{huang2026stgcn}} & proxy critical-node & No & \Wenchuan{no verified executable implementation} \\
ILGR \Wenchuan{\citep{munikoti2022ilgr}} & proxy critical-node & No & \Wenchuan{public \href{https://github.com/saimunikoti/GraphNeuralNetwork-Resilience-ComplexNetworks}{implementation} identified but not integrated} \\
\bottomrule
\end{tabular}
\end{table}

\section{\Wenchuan{Further Discussion}}
\label{app:further-discussion}

\subsection{\Wenchuan{Implications for Learning-based Network Dismantling}}

\Wenchuan{The comparison does not indicate that learning is unsuitable for network dismantling. Instead, it shows that current learning-based methods have not yet established a consistent advantage over mature non-learning approaches. Several learning-based methods obtain competitive results on the networks they complete, but this advantage often weakens when completion coverage and computational cost are considered. FINDER illustrates a different trade-off: it completes all 185 networks, but does not achieve the best dismantling quality. These observations suggest that effectiveness, applicability, and efficiency should be considered together when developing learning-based dismantling methods.}

\Wenchuan{The differences among evaluation metrics also have direct implications for model design. A method trained to minimize the removal set required to reach a prescribed fragmentation level is not necessarily optimized to suppress connectivity throughout the removal trajectory. Likewise, a sharp structural transition does not by itself imply a better dismantling solution. Future learning-based methods should therefore state the intended intervention objective explicitly and evaluate whether the training objective is consistent with that goal. Multi-objective formulations may be useful when both the final intervention cost and the intermediate connectivity of the network are important.}

\Wenchuan{Network scale presents another important challenge. Strong performance on small and medium-sized networks does not guarantee that a method can be applied to larger networks, particularly when inference requires repeated graph processing, graph-specific adaptation, or additional search. Future research should report completion coverage across network sizes and examine which components of the method limit scalability. Improving coverage should not be treated only as an implementation issue, because a method that cannot return a valid solution has limited value in practical dismantling tasks.}

\subsection{\Wenchuan{Scope of the Empirical Evidence}}

\Wenchuan{\textsc{NetBreaker} focuses on static node removal from undirected, unweighted simple graphs. Directed, weighted, temporal, and multilayer networks are outside the present scope. The benchmark also evaluates naturally occurring differences among the collected networks rather than performing controlled interventions on individual graph properties. The domain- and scale-specific results should therefore be interpreted as empirical comparisons within the assembled dataset collection, not as causal claims about how a particular graph property determines method performance.}

\Wenchuan{The benchmark evaluates complete and reproducible methods in the forms made available by their authors. The included learning-based methods differ in training data, released checkpoints, optimization procedures, and post-processing strategies. We retain these differences because they are part of the methods that can currently be reproduced and used. The results consequently compare deployable method pipelines rather than neural architectures retrained under an identical optimization budget. A controlled comparison of learning architectures would answer a related but different research question.}

\subsection{\Wenchuan{Limitations and Benchmark Extensions}}

\Wenchuan{The reported computational costs are observations under the execution protocol used in this study. The evaluated methods rely on heterogeneous CPU and GPU implementations, and their recorded running times may include different combinations of inference, graph-specific adaptation, search, and post-processing. These results support practical comparisons under the stated protocol, but should not be interpreted as hardware-independent measures of algorithmic complexity. Peak host-memory measurements also do not capture all device-memory consumption. Runtime and memory statistics over successful runs must therefore be interpreted together with completion coverage.}

\Wenchuan{Several extensions would broaden the scope of the benchmark. Controlled profiling across hardware environments would provide a more detailed account of computational cost. Retraining selected learning-based methods under a common data and optimization budget could separate architectural capability from differences in existing implementations. The dataset collection could also be extended to directed, weighted, temporal, and multilayer networks, while controlled synthetic networks could be used to study the effect of individual structural properties. These extensions complement the present benchmark and provide directions for examining aspects of network dismantling that are not covered by the current setting.}

\section{\Wenchuan{Additional Results}}
\label{app:additional}

\subsection{\Wenchuan{Detailed Domain- and Scale-specific Results}}

\Wenchuan{Table~\ref{tab:category_domain_scale} complements the method-level ranges in Figure~\ref{fig:domain-scale} with fixed method-family representatives selected once from the overall target-based ranking.}

\begin{table*}[t]
\caption{\Wenchuan{Sensitivity of fixed method-family representatives to network domain and scale. For each family, the representative is selected once by its overall $\rho_{0.10}$ failure-aware rank and then held fixed across all conditions. Entries are normalized failure-aware ranks (lower is better). Gap columns quantify condition sensitivity. Bold marks the best representative in each condition.}}
\label{tab:category_domain_scale}
\centering
\scriptsize
\setlength{\tabcolsep}{2.2pt}
\resizebox{\textwidth}{!}{%
\begin{tabular}{l cccccccc c cccc c}
\toprule
& \multicolumn{8}{c}{\Wenchuan{Network domain}} & & \multicolumn{4}{c}{\Wenchuan{Number of nodes}} & \\
\cmidrule(lr){2-9}\cmidrule(lr){11-14}
\Wenchuan{Method family (fixed representative)} & \Wenchuan{Bio/Eco} & \Wenchuan{Econ/Fin} & \Wenchuan{Info/Media} & \Wenchuan{Science} & \Wenchuan{Security} & \Wenchuan{Social} & \Wenchuan{Tech/Comm} & \Wenchuan{Transport} & \Wenchuan{$\Delta_{\mathrm{domain}}$} & \Wenchuan{$<1$K} & \Wenchuan{1K--10K} & \Wenchuan{10K--100K} & \Wenchuan{$\geq100$K} & \Wenchuan{$\Delta_{\mathrm{scale}}$} \\
\midrule
\Wenchuan{Random (Random)} & \Wenchuan{0.959} & \Wenchuan{0.891} & \Wenchuan{0.865} & \Wenchuan{0.849} & \Wenchuan{0.946} & \Wenchuan{0.786} & \Wenchuan{0.873} & \Wenchuan{0.897} & \Wenchuan{0.173} & \Wenchuan{0.966} & \Wenchuan{0.965} & \Wenchuan{0.825} & \Wenchuan{0.434} & \Wenchuan{0.531} \\
\Wenchuan{Centrality (Adaptive Betweenness)} & \Wenchuan{0.146} & \Wenchuan{0.230} & \Wenchuan{0.297} & \Wenchuan{0.606} & \Wenchuan{0.281} & \Wenchuan{0.398} & \Wenchuan{0.353} & \Wenchuan{\textbf{0.135}} & \Wenchuan{0.471} & \Wenchuan{\textbf{0.153}} & \Wenchuan{\textbf{0.092}} & \Wenchuan{0.675} & \Wenchuan{0.734} & \Wenchuan{0.643} \\
\Wenchuan{Approx./Optimization (BPD)} & \Wenchuan{\textbf{0.121}} & \Wenchuan{\textbf{0.086}} & \Wenchuan{\textbf{0.099}} & \Wenchuan{\textbf{0.075}} & \Wenchuan{\textbf{0.101}} & \Wenchuan{\textbf{0.146}} & \Wenchuan{\textbf{0.102}} & \Wenchuan{0.157} & \Wenchuan{0.082} & \Wenchuan{0.172} & \Wenchuan{0.104} & \Wenchuan{\textbf{0.064}} & \Wenchuan{\textbf{0.147}} & \Wenchuan{0.107} \\
\Wenchuan{Supervised (CoreGDM)} & \Wenchuan{0.223} & \Wenchuan{0.299} & \Wenchuan{0.279} & \Wenchuan{0.265} & \Wenchuan{0.234} & \Wenchuan{0.293} & \Wenchuan{0.209} & \Wenchuan{0.217} & \Wenchuan{0.089} & \Wenchuan{0.213} & \Wenchuan{0.224} & \Wenchuan{0.190} & \Wenchuan{0.535} & \Wenchuan{0.345} \\
\Wenchuan{Self-/Unsupervised (Stable-AGDM+R)} & \Wenchuan{0.207} & \Wenchuan{0.368} & \Wenchuan{0.263} & \Wenchuan{0.211} & \Wenchuan{0.377} & \Wenchuan{0.286} & \Wenchuan{0.192} & \Wenchuan{0.240} & \Wenchuan{0.185} & \Wenchuan{0.243} & \Wenchuan{0.203} & \Wenchuan{0.162} & \Wenchuan{0.543} & \Wenchuan{0.381} \\
\Wenchuan{Reinforcement (FINDER)} & \Wenchuan{0.494} & \Wenchuan{0.313} & \Wenchuan{0.441} & \Wenchuan{0.501} & \Wenchuan{0.335} & \Wenchuan{0.381} & \Wenchuan{0.454} & \Wenchuan{0.489} & \Wenchuan{0.188} & \Wenchuan{0.457} & \Wenchuan{0.489} & \Wenchuan{0.440} & \Wenchuan{0.242} & \Wenchuan{0.247} \\
\Wenchuan{Generative Search (LLM4CN)} & \Wenchuan{0.704} & \Wenchuan{0.701} & \Wenchuan{0.747} & \Wenchuan{0.828} & \Wenchuan{0.670} & \Wenchuan{0.776} & \Wenchuan{0.895} & \Wenchuan{0.745} & \Wenchuan{0.225} & \Wenchuan{0.713} & \Wenchuan{0.737} & \Wenchuan{0.936} & \Wenchuan{0.734} & \Wenchuan{0.223} \\
\Wenchuan{Proxy Critical-node (AGNN)} & \Wenchuan{0.860} & \Wenchuan{0.833} & \Wenchuan{0.841} & \Wenchuan{0.780} & \Wenchuan{0.837} & \Wenchuan{0.833} & \Wenchuan{0.841} & \Wenchuan{0.858} & \Wenchuan{0.080} & \Wenchuan{0.868} & \Wenchuan{0.871} & \Wenchuan{0.802} & \Wenchuan{0.734} & \Wenchuan{0.137} \\
\bottomrule
\end{tabular}%
}
\end{table*}

\subsection{\Wenchuan{Complete Metric-dependent Rank Profiles}}

\begin{figure*}[t]
	\centering
	\includegraphics[width=0.94\textwidth]{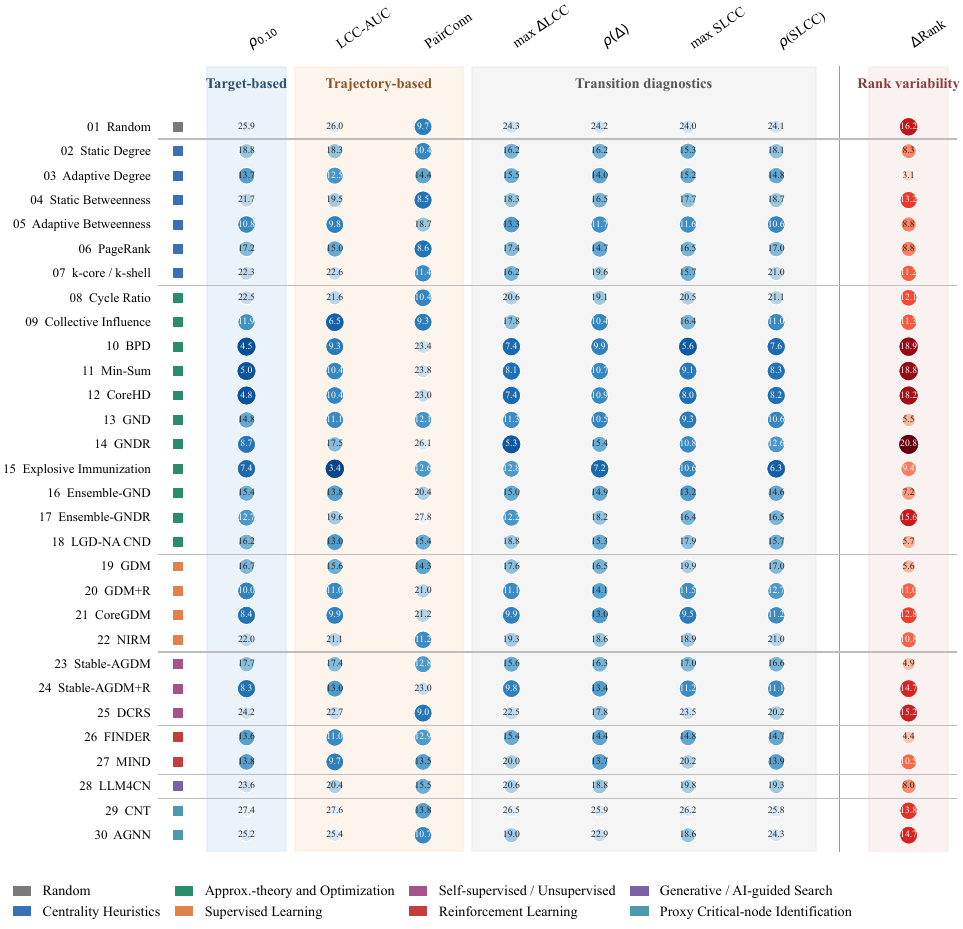}
	\caption{\Wenchuan{Complete failure-aware rank profiles across evaluation perspectives. Darker and larger blue circles indicate better mean ranks; the red column reports the range between each method's best and worst rank. Transition magnitudes characterize the fragmentation mechanism rather than an independent overall objective.}}
	\label{fig:metric-ranks-appendix}
\end{figure*}

\Wenchuan{Table~\ref{tab:metric_top5} summarizes the five highest-ranked methods under the target criterion and two trajectory summaries. Figure~\ref{fig:metric-ranks-appendix} extends this comparison to all methods and all seven evaluation views. The target and trajectory criteria measure dismantling quality, whereas the transition diagnostics characterize when and how the residual network fragments.}

\begin{table*}[t]
\caption{\Wenchuan{Top-5 methods under the target-based criterion and the two trajectory-based criteria. Entries report failure-aware mean ranks over 185 networks; lower is better.}}
\label{tab:metric_top5}
\centering
\scriptsize
\setlength{\tabcolsep}{4.0pt}
\begin{tabular}{c lc lc lc}
\toprule
& \multicolumn{2}{c}{\Wenchuan{Target-based}} & \multicolumn{4}{c}{\Wenchuan{Trajectory-based}} \\
\cmidrule(lr){2-3}\cmidrule(lr){4-7}
\Wenchuan{Rank} & \multicolumn{2}{c}{\Wenchuan{$\rho_{0.10}$}} & \multicolumn{2}{c}{\Wenchuan{LCC-AUC}} & \multicolumn{2}{c}{\Wenchuan{PairConn}} \\
& \Wenchuan{Method} & \Wenchuan{Mean rank $\downarrow$} & \Wenchuan{Method} & \Wenchuan{Mean rank $\downarrow$} & \Wenchuan{Method} & \Wenchuan{Mean rank $\downarrow$} \\
\midrule
\Wenchuan{1} & \Wenchuan{\textbf{BPD}} & \Wenchuan{\textbf{4.45}} & \Wenchuan{\textbf{Explosive Immunization}} & \Wenchuan{\textbf{3.37}} & \Wenchuan{\textbf{Static Betweenness}} & \Wenchuan{\textbf{8.46}} \\
\Wenchuan{2} & \Wenchuan{CoreHD} & \Wenchuan{4.83} & \Wenchuan{Collective Influence} & \Wenchuan{6.45} & \Wenchuan{PageRank} & \Wenchuan{8.59} \\
\Wenchuan{3} & \Wenchuan{Min-Sum} & \Wenchuan{4.97} & \Wenchuan{BPD} & \Wenchuan{9.33} & \Wenchuan{DCRS} & \Wenchuan{9.01} \\
\Wenchuan{4} & \Wenchuan{Explosive Immunization} & \Wenchuan{7.41} & \Wenchuan{MIND} & \Wenchuan{9.70} & \Wenchuan{Collective Influence} & \Wenchuan{9.34} \\
\Wenchuan{5} & \Wenchuan{Stable-AGDM+R} & \Wenchuan{8.26} & \Wenchuan{Adaptive Betweenness} & \Wenchuan{9.84} & \Wenchuan{Random} & \Wenchuan{9.74} \\
\bottomrule
\end{tabular}
\end{table*}

\subsection{\Wenchuan{Trajectory-based Quality--Cost Trade-offs}}

\begin{figure*}[t]
	\centering
	\includegraphics[width=0.88\textwidth]{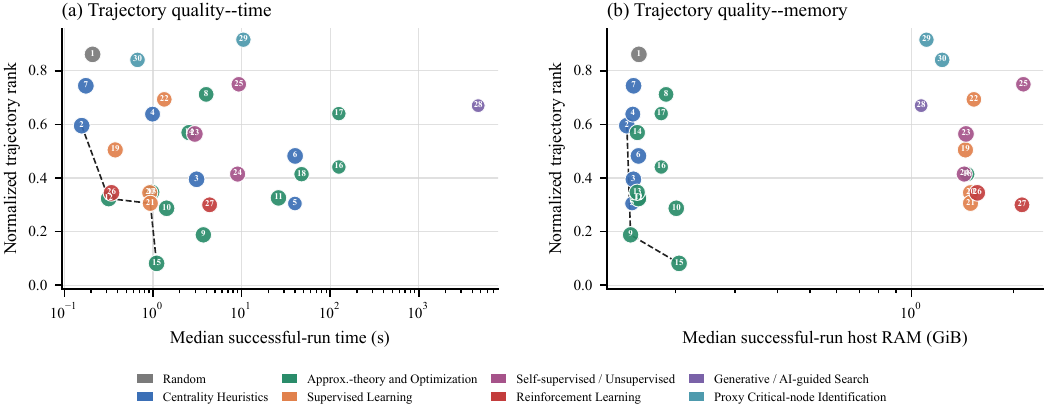}
	\caption{\Wenchuan{Observed trajectory-quality cost trade-offs over 185 networks. Panels (a) and (b) compare the failure-aware LCC-AUC rank with median successful-run time and peak host memory, respectively. Lower and further left is better; dashed lines mark non-dominated observations, marker area encodes completion coverage, and method IDs refer to Table~\ref{tab:main_results_full}.}}
	\label{fig:trajectory-cost-appendix}
\end{figure*}

\Wenchuan{Figure~\ref{fig:trajectory-cost-appendix} complements the target-based cost analysis in Figure~\ref{fig:cost-scaling} by evaluating the connectivity retained throughout the removal trajectory.}

\end{document}